\documentclass[11pt]{article}

\usepackage[final]{acl}

\usepackage{times}
\usepackage{latexsym}

\usepackage[T1]{fontenc}

\usepackage[utf8]{inputenc}

\usepackage{microtype}

\usepackage{inconsolata}

\usepackage{graphicx}
\usepackage{enumitem}
\usepackage[most]{tcolorbox}
\tcbuselibrary{skins}
\usepackage{xcolor}
\usepackage{algorithm}
\usepackage{algorithmic}
\usepackage{amsmath}
\usepackage{amssymb}
\usepackage{xspace}
\usepackage{booktabs}
\usepackage{multirow}
\usepackage{geometry}
\usepackage{todonotes}
\usepackage{makecell}
\usepackage{url}
\usepackage{graphicx}
\usepackage[table]{xcolor}
\definecolor{tagcolor}{RGB}{128, 128, 128}
\definecolor{headergray}{RGB}{240, 240, 240}

\definecolor{failbg}{RGB}{255, 235, 235} 
\definecolor{failframe}{RGB}{200, 100, 100}
\definecolor{passbg}{RGB}{235, 255, 235} 
\definecolor{passframe}{RGB}{100, 200, 100}
\definecolor{codeblue}{RGB}{0, 0, 150}
\definecolor{hallucinationred}{RGB}{220, 0, 0}
\definecolor{hallucinationbg}{RGB}{255, 240, 240}
\definecolor{basecolor}{RGB}{255, 255, 204}
\definecolor{toolcolor}{RGB}{255, 234, 230}
\definecolor{spectracolor}{RGB}{224, 255, 224}
\definecolor{ablatecolor}{RGB}{204, 232, 255}
\definecolor{spectraheading}{RGB}{200, 245, 200}

\newtcolorbox{mainbox}[1]{
  enhanced,
  colback=white,
  colframe=gray!80,
  fonttitle=\bfseries,
  title={#1},
  boxrule=0.3mm,
  drop shadow,
  left=1mm, right=1mm, top=1mm, bottom=2mm 
}

\newtcolorbox{subbox}[3]{
  enhanced,
  colback=#1,
  colframe=#2,
  boxrule=0.3mm,
  arc=1mm,
  left=1.5mm, right=1.5mm, top=1mm, bottom=1mm, 
  fonttitle=\bfseries\footnotesize,
  title={#3},
  coltitle=black,
  attach boxed title to top left={yshift=-2mm, xshift=2mm},
  boxed title style={colback=#1, frame hidden}
}

\newtcolorbox{baselinebox}{
  enhanced,
  colback=failbg,
  colframe=failframe,
  boxrule=0.2mm,
  arc=2mm,
  left=2mm, right=2mm, top=2mm, bottom=2mm,
  title={\textbf{$\times$ Baseline Agent}},
  coltitle=black,
  attach boxed title to top left={yshift=-1mm, xshift=1mm},
  boxed title style={colback=failbg, frame hidden}
}

\newtcolorbox{spectrabox}{
  enhanced,
  colback=passbg,
  colframe=passframe,
  boxrule=0.2mm,
  arc=2mm,
  left=2mm, right=2mm, top=2mm, bottom=2mm,
  title={\textbf{$\checkmark$ SPECTRA (Ours)}},
  coltitle=black,
  attach boxed title to top left={yshift=-2mm, xshift=2mm},
  boxed title style={colback=passbg, frame hidden}
}

\newcommand{\toolcall}[1]{{\small\ttfamily\color{codeblue} #1}}
\newcommand{\hallucination}[1]{%
  \begingroup
  \setlength{\fboxsep}{1pt}%
  \colorbox{hallucinationbg}{\scriptsize\ttfamily\color{hallucinationred}\textbf{#1}}%
  \endgroup
}

\DeclareCaptionFormat{coloredheader}{%
  \colorbox{headergray}{\parbox{\dimexpr\linewidth-2\fboxsep\relax}{#1#2#3}}%
}
\newcommand{\drop}[1]{_{\downarrow \textcolor{red}{#1}}}
\newcommand{\uparw}[1]{_{\uparrow \textcolor{gray}{#1}}}
\newcommand{\spectra}{\texttt{SPECTRA}\xspace}
\newcommand\blfootnote[1]{%
  \begingroup
  \renewcommand\thefootnote{}\footnote{#1}%
  \addtocounter{footnote}{-1}%
  \endgroup
}
\newcommand{\spectrafull}{\textbf{Self-supervised Perception Enabled by Cascaded Tool Rollout Alignment (\spectra)}}

\title{\textit{Waking Up Blind}: Cold-Start Optimization of Supervision-Free Agentic Trajectories for Grounded Visual Perception}

\author{\textbf{Ashutosh Bajpai\textsuperscript{1,*}},  \textbf{Tamal Majumder\textsuperscript{1,*}},  \textbf{Akshay Nambi\textsuperscript{3}},  \textbf{Tanmoy Chakraborty\textsuperscript{1,2}}\\
\\
\textsuperscript{1}Indian Institute of Technology Delhi, India \\    
\textsuperscript{2}Indian Institute of Technology Abu Dhabi, UAE \\ 
\textsuperscript{3}Microsoft Research, India \\    
\textit{\{eez228482,tanchak\}@ee.iitd.ac.in}, \textit{987tamal@gmail.com}, \\
\textit{akshayn@microsoft.com}}    

\begin{document}
\maketitle
\begin{abstract}
Small Vision-Language Models (SVLMs) are efficient task controllers but often suffer from visual brittleness and poor tool orchestration. They typically require expensive supervised trajectory tuning to mitigate these deficits. In this work, we propose \spectrafull, a \textit{supervision-free framework} that bootstraps agentic capabilities via Coldstart Reinforcement Learning for SVLMs. \spectra\ enforces \textit{Soft Structured Multi-turn Rollouts}, a topological constraint that directs agents to explicitly sequence tool-derived evidence before synthesis, effectively grounding reasoning in visual observations. We employ a multi-objective reward signal that simultaneously maximizes task correctness, rollout structure, and tool utility, enabling agent to self-discover robust behaviors without human preference labels. We further introduce \textbf{Tool Instrumental Utility (TIU)}, a novel metric to quantify tool efficacy in the absence of ground truth. Extensive evaluations across composite and out-of-distribution (MMMU-Pro) benchmarks demonstrate that \spectra\ boosts agentic trajectories, improving task accuracy by up to 5\% and tool efficiency by 9\%, enabling more efficient multimodal agents that learn effectively from environmental interaction alone. \blfootnote{\textsuperscript{*}Equal Contribution.} 
\end{abstract}
\begin{figure*}[!t]
 \centering
\includegraphics[width=1\textwidth]{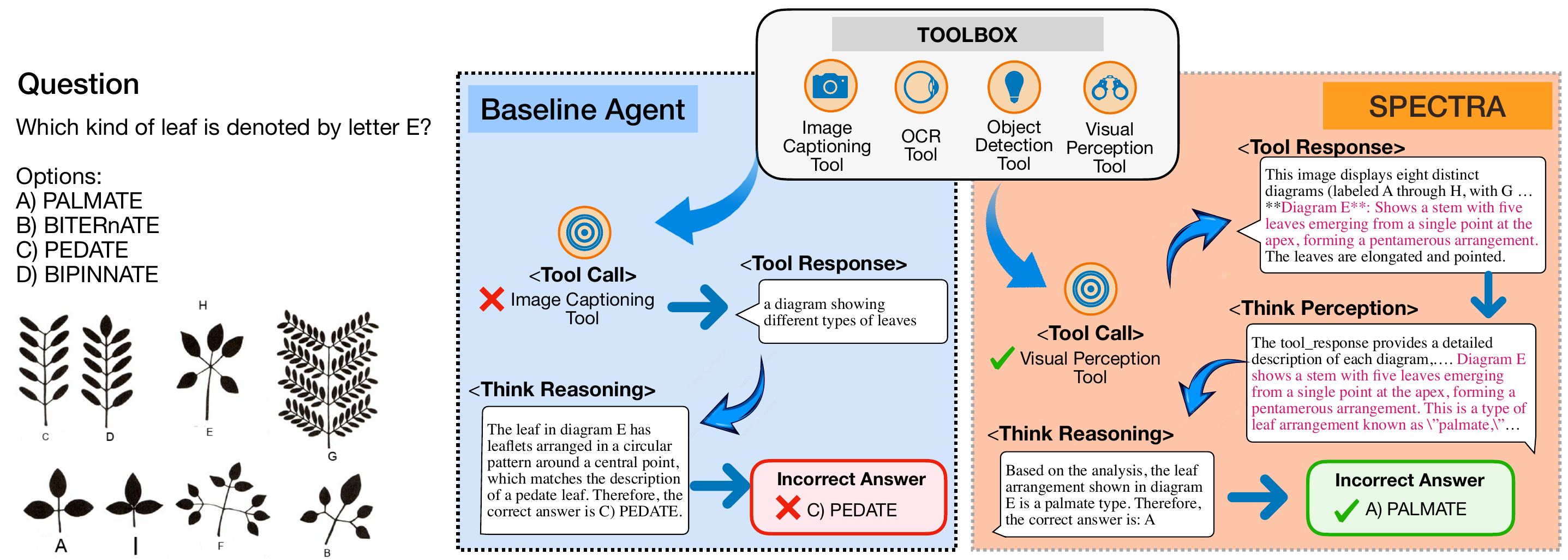}
  \caption{Comparison of the Qwen-2.5-VL-7B based \spectra\ with a baseline agent, showcasing its efficient tool utilization and visually-guided structured approach, resulting in accurate outcomes.} 
  \label{fig:main_daigram}
\end{figure*}
\section{Introduction}
\textbf{SVLMs As Autonomous Controllers.} Foundation models are rapidly transitioning from passive reasoning to agentic systems that perceive, reason, and act across multimodal, multi-step tasks \cite{bommasani2022opportunitiesrisksfoundationmodels,yang2023foundationmodelsdecisionmaking}. Within this shift, small vision-language models (SVLMs) such as Qwen2.5-VL \cite{bai2025qwen25vltechnicalreport} are attractive controllers due to favorable latency and deployment cost, yet they lag larger VLM controllers in long-horizon reasoning and suffer from weak fine-grained visual perception and inefficient tool orchestration \cite{yao2024minicpmvgpt4vlevelmllm,wang2024qwen2vlenhancingvisionlanguagemodels,Tong2024EyesWS,guan2024hallusionbenchadvanceddiagnosticsuite}. Prior work shows that stepwise reasoning with tool calls can be scaffolded by prompting frameworks like ReAct \cite{yao2023reactsynergizingreasoningacting} and that VLM-driven agents can outperform LLM-only controllers when visual context determines which tools to call \cite{gao2025multimodal,Zheng2024GPT4VisionIA}. Despite these advances, SVLMs still struggle to decompose multimodal problems into reliable stepwise trajectories that balance perception, reasoning, and tool usage under data scarcity \cite{Li2025TextureOS}. 
\paragraph{Recent advancements in agentic trajectories optimization.} 
Two approaches work to close the aforementioned gap. First, trajectory tuning with verifiable synthetic tool-use data improves VLM controllers' stepwise tool reasoning (LLaVA-Plus \cite{liu2023llavapluslearningusetools}, MLLM-Tool \cite{wang2025mllmtoolmultimodallargelanguage}), with T3-Agent introducing MM-Traj, 
demonstrating ~$20\%$ gains on GTA \cite{wang2024gta} via ReAct-style trajectory tuning of MiniCPM-V-8.5B and Qwen2-VL-7B \cite{gao2025multimodal}. Larger-scale resources extend this direction, with evaluators increasingly assessing trajectory quality rather than focusing solely on final answers \cite{anonymous2025beyond}. Second, Reinforcement Learning (RL) for tool-augmented agents emphasizes sample efficiency and adaptive perception (e.g., Tool-R1 \cite{zhang2025tonguiinternetscaletrajectoriesmultimodal}), while tool integration and memory mechanisms (e.g., AutoTool \cite{zou2025autotooldynamictoolselection}, ToolMem \cite{xiao2025toolmemenhancingmultimodalagents}) reduce redundant calls and better align tools with task demands.
\paragraph{Limitations of trajectory tuning.} Despite aforementioned advances, most systems either (i) tune for tool-use reasoning without directly refining visual perception via structured rollouts, or (ii) apply supervised data-driven mechanisms to align tool efficiency, visual evidence acquisition, and stepwise reasoning, especially relying on expensive synthetically generated data.

We investigate how tool and structured rollouts-mediated visual perception refinement mechanisms can enhance SVLMs’ agentic capabilities, enabling better task decomposition into stepwise trajectories and bridging task-specific tool alignments in absence of supervised tool trajectories. We pose two research questions:
\begin{itemize} [itemsep=1pt,left=0pt]
    \item \textbf{RQ1:} How can SVLMs’ agentic capabilities be augmented by improving tool-calling efficiency via coldstart RL, i.e., learning from environment signals without human preference models?
    \item \textbf{RQ2:} How can improving visual perception through soft structured rollouts address SVLMs’ weak visual perception?
\end{itemize}
To address these questions, we introduce the \spectrafull. Unlike traditional approaches dependent on supervised demonstrations, \spectra\ optimizes an SVLM policy directly from environmental interactions using a multi-objective reward signal that simultaneously maximizes task correctness, rollout structure, and instrumental tool utility via Group Relative Policy Optimization (GRPO) strategy. We enforce a hierarchical reasoning topology,  \textit{Soft Structured Multi-turn Rollouts}, which compels the agent to explicitly sequence tool selection, observation integration, and perceptual synthesis before answering. By employing a multi-objective reward signal that balances task correctness with structural integrity and tool diversity, \spectra\ enables the model to self-discover robust agentic behaviors and ground its visual perception in tool-derived evidence, effectively bypassing the need for human preferred tool trajectories or reasoning traces often distilled from large models. Moreover, Tool Accuracy \cite{gao2025multimodal} is commonly used to evaluate tool trajectories. However, limited attention has been given to evaluating tool correctness or efficiency in absence of supervised trajectories. 

To address this limitation, we propose a novel metric, \textbf{Tool Instrumental Utility (TIU)}, which combines tool reliability, alignment, and selectivity to offer a comprehensive measure. We release \spectra in 7B and 3B variants, both utilizing the Qwen-2.5-VL. \spectra\ boosts agentic trajectories, improving task accuracy by up to 5\% and tool efficiency by 9\% over closest baseline across four multimodal benchmarks. An illustration of \spectra\ is presented in Figure \ref{fig:main_daigram}.

The primary contributions of this work are summarized as follows\footnote{Source code and dataset are available at \url{https://github.com/ab-iitd/spectra}}:
\begin{itemize}
[noitemsep,nolistsep,topsep=2pt,leftmargin=1em]
    \item We introduce soft structured multi-turn topological (rollout) constraints to enforce focused and grounded stepwise visual reasoning.
    \item We propose \spectra, a novel supervision-free agentic policy optimization for tool trajectories and visual reasoning traces.
    \item We propose TIU, a novel metric to measure tool efficiency in the absence of supervised tool trajectory preferences.
    \item Empirical studies with qualitative analysis demonstrate that \spectra\ outperform strong baselines in enhancing both task accuracy and agent trajectories.
\end{itemize}

\section{Related Works}
\textbf{Multimodal agents and controllers.} Early multimodal agents  utilized large language models (LLMs) to plan and execute tool calls via textual interfaces, often employing ReAct-style reasoning or prompting frameworks \cite{surís2023vipergptvisualinferencepython,gupta2022visualprogrammingcompositionalvisual,shen2023hugginggpt,yao2023reactsynergizingreasoningacting,yang2023foundationmodelsdecisionmaking}. However, relying solely on textual queries can hinder tool selection when visual context is critical \cite{Fan2024VideoAgentAM,wu2023visualchatgpttalkingdrawing,trivedi2024appworldcontrollableworldapps}. To mitigate this, VLM-driven agents integrate the controller within a multimodal model, allowing direct conditioning on visual inputs to improve efficiency in vision-centric tasks \cite{trivedi2024appworldcontrollableworldapps,zheng2024gpt4visiongeneralistwebagent,wang2024genartist,zheng2024steveeye}. Addressing the weaker reasoning capabilities of smaller VLMs, recent approaches finetune models on tool-usage trajectories \cite{liu2023llavapluslearningusetools,wang2025mllmtoolmultimodallargelanguage}. Notably, T3-Agent \cite{gao2025multimodal} leverages automatically generated trajectories
, demonstrating substantial improvements on benchmarks like GTA and GAIA \cite{mialon2023gaiabenchmarkgeneralai}.

\textbf{RL for decision-making VLM agents.} Interactive RL offers a pathway to teach agents multi-step decision-making using environmental feedback. While RLHF aligns LLMs with human preferences \cite{ouyang2022traininglanguagemodelsfollow,Stiennon2020LearningTS}, task-specific RL has proven effective in text-only domains \cite{ramamurthy2023reinforcementlearningnotnatural,zhou2024archertraininglanguagemodel}. Extending this to VLMs, RL4VLM \cite{zhai2024finetuninglargevisionlanguagemodels} fine-tunes a 7B model using environmental rewards on tasks requiring visual and linguistic understanding. \citet{zhang2025toolr1sampleefficientreinforcementlearning} expanded on this to introduce sample-efficient RL for tool use. Despite training instability challenges in GRPO \cite{deng2025grpocollapsesearchr1lazy}, these studies confirm that RL can effectively equip SVLM controllers \cite{schulman2017proximalpolicyoptimizationalgorithms,haarnoja2018softactorcriticoffpolicymaximum}.

\textbf{Tool-usage datasets and trajectory tuning.} Agents depend on high-quality tool-usage data capturing queries, multimodal files, thoughts, actions, and observations. Early tool datasets, such as APIBank \cite{li2023apibankcomprehensivebenchmarktoolaugmented}; ToolAlpaca \cite{tang2023toolalpacageneralizedtoollearning} ToolBench \cite{qin2024toolllm}; AnyTool \cite{du2024anytoolselfreflectivehierarchicalagents}; AgentOhana \cite{zhang2024agentohanadesignunifieddata}; APIGen \cite{liu2024apigen}; AgentInstruct \cite{mitra2024agentinstructgenerativeteachingagentic},  focus primarily on text tools, with limited multimodal coverage. Benchmarks such as OSWorld \cite{xie2024osworld}, MMInA \cite{tian2025mminabenchmarkingmultihopmultimodal}, AgentBench \cite{liu2025agentbenchevaluatingllmsagents}, and AgentGym \cite{xi2024agentgymevolvinglargelanguage} incorporate multimodality, but datasets for training tool-augmented VLMs at scale are scarce. MLLM-Tool \cite{wang2025mllmtoolmultimodallargelanguage} and T3-Agent \cite{gao2025multimodal} overcome the aforementioned challenge by advancing tool-trajectories data generation methods. Current methods for improving perception and tool efficiency in multimodal agents predominantly rely on supervised signals from curated tool trajectories and structured reasoning traces. However, acquiring such high-quality supervision at scale is costly and limits generalization. To overcome these challenges, we propose \spectra, a novel method that leverages structured rollout and unsupervised cold-start RL to enhance agentic trajectories without reliance on expensive ground-truth annotations.

\section{Dataset and Benchmarks}
To foster robust agentic reasoning in SVLMs, we consider four public distinct domains datasets: AI2D \cite{kembhavi2016diagramworthdozenimages}, TQA \cite{kim2019textbookquestionansweringmultimodal}, ScienceQA \cite{lu2022learnexplainmultimodalreasoning}, and OK-VQA \cite{marino2019okvqavisualquestionanswering}. These benchmarks are selected to ensure broad coverage of multi-modal capabilities, ranging from diagrammatic parsing and textbook-length comprehension to external knowledge retrieval and chain-of-thought reasoning. We use 1,000 training and 200 test samples per dataset (4,000 and 800 total) to ensure sample-efficient learning. For rigorous out-of-distribution evaluation, we employ single-image MMMU-Pro \cite{yue2025mmmuprorobustmultidisciplinemultimodal}, a challenging 30-subject university-level benchmark for testing integrated visual-textual reasoning (Appendix \ref{app:benchmarks}).

\begin{figure*}[!t]
 \centering
\includegraphics[width=1\textwidth]{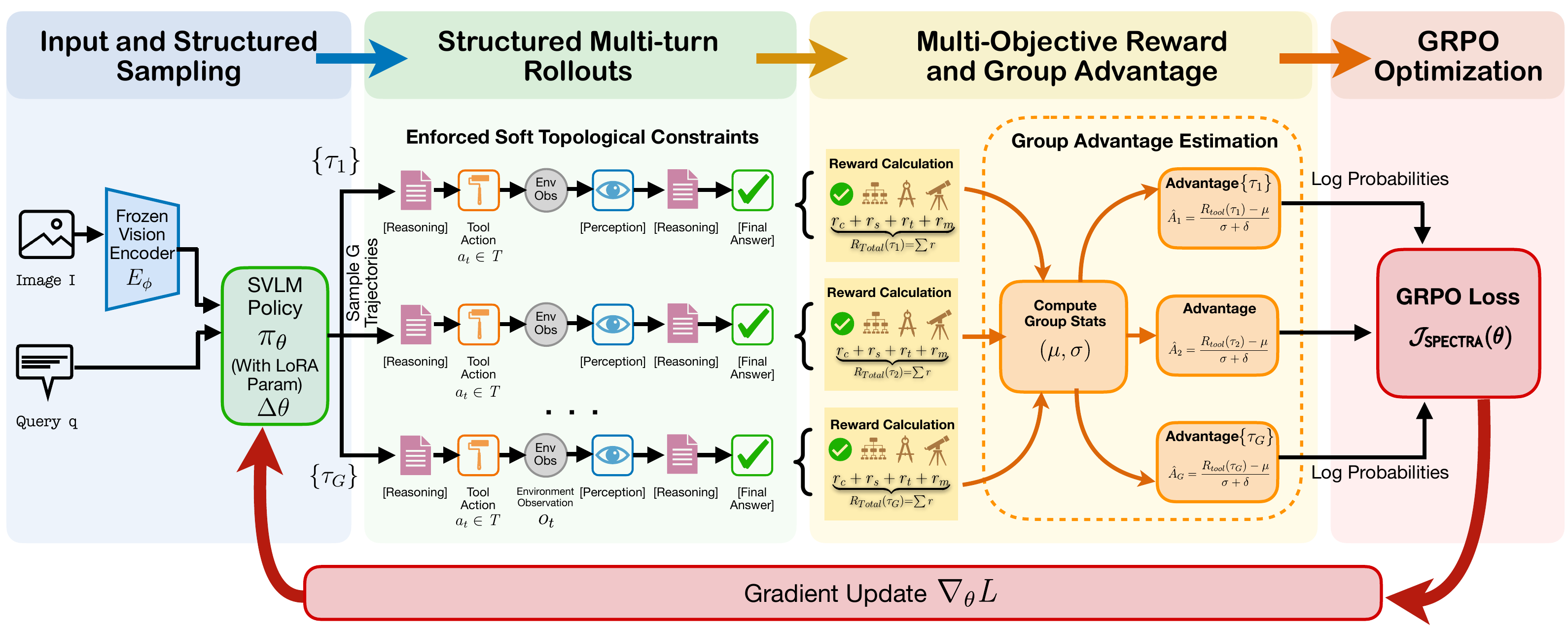}
  \caption{Overview of \spectra. The pipeline samples structured multi-turn trajectories from the SVLM policy, aggregates multi-objective rewards to compute group-relative advantages, and iteratively optimizes the policy parameters via GRPO-based objective, where $r_c:R_{corr}, r_s:R_{struct}, r_t:R_{tool}, r_m:R_{term}$ are reward components for task accuracy, structural integrity, tool efficiency, and terminal delimitation, respectively.} 
  \label{fig:spectra_formulation}
\end{figure*}

\section{Methodology}
\label{sec:methodology}
We introduce \spectrafull\ (Figure \ref{fig:spectra_formulation}), to augment SVLMs tool-use and visual grounding by using domain-specialized GRPO to learn directly from environmental signals, bypassing the need for supervised demonstrations.
\subsection{Problem Formulation}
We formulate the agentic reasoning task similar to a Partial Observation Markov Decision Process (POMDP) \cite{ASTROM1965174}. Let the multimodal input space be defined by tuples $(I, q) \sim \mathcal{D}$, where $I$ is the visual context (image) and $q$ is the natural language query.

\paragraph{Policy Architecture.}
The agent is parameterized by an SVLM policy $\pi_\theta$. The architecture utilizes a frozen vision encoder $E_\phi$ and a trainable LLM decoder adapted via Low-Rank Adaptation (LoRA). We optimize only the adapter parameters $\Delta \theta$, keeping the base vision-language alignment fixed to preserve general knowledge. The policy output at time step $t$ is given by:
\vspace{-0.1em}
\begin{equation}
    \pi_\theta(a_t | s_t) = \text{Softmax}\left( W_{\text{frozen}} h_t + B A h_t \right)
\end{equation}
where $W_{\text{frozen}}$ represents the fixed pre-trained weights, $B, A \in \mathbb{R}^{d \times r}, \mathbb{R}^{r \times d}$ are the low-rank learnable matrices, $h_t$ is latent representation of input, and $a_t$ denotes action at time $t$ given state $s_t$.

\paragraph{Action Space.}
The action space $\mathcal{A}$ is a union of natural language tokens and a discrete set of tool primitives ${T} = \{ T_{\text{cap}}, T_{\text{det}}, T_{\text{ocr}}, T_{\text{vp}} \}$, corresponding to Image Captioning, Object Detection, OCR, and Visual Perception respectively. The detailed description of tools are provided in Table \ref{tab:tooldesc}.
\subsection{Soft Structured Multi-turn Rollouts}

Drawing inspiration from \citet{xue2025simpletirendtoendreinforcementlearning}, we address the ``blindness'' often observed in SVLMs by enforcing a hierarchical reasoning structure. An optimal trajectory $\tau$ must adhere to the following topological sequence:
\begin{equation}
\label{eq:rollout}
\resizebox{0.98\hsize}{!}{%
$\tau = \langle \texttt{reason} \to \texttt{tool} \to \texttt{obs} \to \texttt{percep} \to \texttt{reason} \to \texttt{ans} \rangle$%
}
\end{equation}
This structure forces the model to explicitly reason about \textit{which} tool to use, integrate the tool's raw output (Observation), synthesize that output with visual features (Perception) and re-reason before committing to a final answer (Figure \ref{fig:rollout}).

\subsection{Cold-Start Agentic Optimization}
We introduce a specialized optimization objective that enables the SVLM to self-explore valid tool-usage patterns without human reference trajectories. We employ a GRPO style policy gradient with group‑normalized advantages \cite{shao2024deepseekmathpushinglimitsmathematical}. Unlike standard RLHF, which constrains the policy to a semantic reference, our ``Cold-Start'' setting requires the model to significantly deviate from its initial distribution to discover agentic behaviors. Therefore, we rely on group-relative normalization to stabilize the learning of diverse reasoning paths.

For each multimodal input $(I, q)$, we sample a group of $G$ distinct cognitive rollouts $\{\tau_1, \tau_2, \dots, \tau_G\}$ from the current policy $\pi_{\theta_{\text{old}}}$. These trajectories represent competing reasoning strategies for the same visual stimulus. The GRPO objective for \spectra\ is formulated as:
\vspace{-0.2cm}
\begin{multline}
\small
\mathcal{J}_{\text{\spectra}}(\theta)
=
\mathbb{E}_{(I,q)\sim\mathcal{D},\,\{\tau_i\}_{i=1}^G \sim \pi_{\theta_{\text{old}}}}
\Bigg[
\frac{1}{G} \sum_{i=1}^G
\frac{1}{|\tau_i|} \\\sum_{t=1}^{|\tau_i|}
\small \min \Big(
\rho_{i,t} \,\hat{A}_{i,t},
\operatorname{clip}\big(\rho_{i,t}, 1-\epsilon_{\text{l}}, 1+\epsilon_{\text{h}}\big)\,\hat{A}_{i,t}
\Big)
\Bigg]\\
\small- \psi \, D_{\mathrm{KL}}\!\left(\pi_\theta\;\|\;\pi_{\theta_{\text{ref}}}\right)
\end{multline}
where $\rho_i = \frac{\pi_\theta(\tau_i | I, q,\tau_{i,<t})}{\pi_{\theta_{\text{old}}}(\tau_i | I, q,\tau_{i,<t})}$ represents the probability ratio. $\hat{A}_i$ is the \textbf{Trajectory Advantage}, which standardizes the dense multi-objective reward $R_{\text{total}}$ against the group's baseline performance:
\vspace{-0.2cm}
\begin{equation}
    \hat{A}_i = \frac{R_{\text{total}}(\tau_i) - \mu(\{R_{\text{total}}(\tau_j)\}_{j=1}^G)}{\sigma(\{R_{\text{total}}(\tau_j)\}_{j=1}^G) + \delta}
\end{equation}
Here, $\mu$ and $\sigma$ act as a dynamic, input-specific baseline computed over sampled group, effectively filtering out the inherent difficulty of the visual query $I$ and isolating the quality of the tool-usage strategy.

\begin{figure}
\begin{center}
\begin{tcolorbox}[
    colback=gray!5, 
    colframe=gray, 
    boxrule=0.3pt, 
    arc=1mm, 
    title=\textbf{Definition: Multi-Turn Structured Rollout ($\tau$)},
    fonttitle=\bfseries\small,
    width=\linewidth
]
\scriptsize
\vspace{-0.1cm}
\textbf{1. Reasoning Phase:} \\
\texttt{\textbf{\textcolor{tagcolor}{<think\_reasoning>}}} 
Analyzes query complexity and plans the next step. 
\texttt{\textbf{\textcolor{tagcolor}{</think\_reasoning>}}}

\vspace{0.4em}
\textbf{2. Tool Execution:} \\
\texttt{\textbf{\textcolor{tagcolor}{<tool\_call>}}} 
Invokes module $T_i \in {T}$ with arguments  
\texttt{\textbf{\textcolor{tagcolor}{</tool\_call>}}}

\vspace{0.4em}
\textbf{3. Observation Integration:} \\
\texttt{\textbf{\textcolor{tagcolor}{<tool\_response>}}} 
Ingests the raw external tool output. 
\texttt{\textbf{\textcolor{tagcolor}{</tool\_response>}}}

\vspace{0.4em}
\textbf{4. Perceptual Re-alignment:} \\
\texttt{\textbf{\textcolor{tagcolor}{<think\_perception>}}} 
Synthesizes tool output with visual features. \texttt{\textbf{\textcolor{tagcolor}{</think\_perception>}}}

\vspace{0.4em}
\begin{center}
    \textit{... [Iterative Process] ...}
\end{center}

\vspace{0.3em}
\textbf{5. Conclusion:} \\
\texttt{\textbf{\textcolor{tagcolor}{<answer>}}} 
Generates the final prediction. 
\texttt{\textbf{\textcolor{tagcolor}{</answer>}}}
\vspace{-0.1cm}
\end{tcolorbox}
\vspace{-0.2cm}
\end{center}
  \caption{Multi-turn structured rollout template.}
  \label{fig:rollout}
\end{figure}

\subsection{Multi-Objective Agentic Reward}

The optimization is driven by $R_{\text{total}}(\tau)$, a scalarized signal that decomposes the agent's behavior into correctness, structural integrity, and tool efficiency:
\vspace{-0.2cm}
\begin{align}\scriptsize
    R_{\text{total}}(\tau) &=  \lambda_1 R_{\text{corr}} +  \lambda_2 R_{\text{struct}} +  \lambda_3R_{\text{tool}} +  \lambda_4R_{\text{term}}
\end{align}
Where $\lambda^*$ denotes the reward component weight.
\paragraph{Task Correctness ($R_{\text{corr}}$).} A sparse signal indicating if the final answer $y_{\text{pred}}$ matches the ground truth $y_{\text{gt}}$:
\vspace{-0.2cm}
\begin{equation}
    R_{\text{corr}}(\tau) = C1 \cdot \mathbb{1}(y_{\text{pred}} = y_{\text{gt}})
\end{equation}
where $C1 \in \mathbb{Z}$ represents the reward magnitude and $\mathbb{1}$ is the binary indicator function. 
\paragraph{Hierarchical Structural Integrity ($R_{\text{struct}}$).} To enforce the \textit{Soft Structured Multi-turn Rollout} (Eq. \ref{eq:rollout}), we gradually penalize deviations from the optimal reasoning topology to ensure other valid paths are not marginalized. We define a hierarchy mapping $\phi(\tau)$, where $\phi=0$ represents a complete agentic loop. The reward decays as the structure degrades from optima:
\vspace{-0.2cm}
\begin{equation}
    R_{\text{struct}}(\tau) = \alpha \cdot \gamma^{\phi(\tau)}
\end{equation}
where $\alpha=2.0$, $\gamma=0.75$, and $\phi(\tau)$ map partial structures (e.g., missing perception) to higher integers. Please refer Appendix \ref{app:reward_struct} for more details.

\paragraph{Tool Utility ($R_{\text{tool}}$).} This component ensures the agent generates executable tool calls and avoids hallucinating non-existent tools. The total tool utility is the sum of a syntactic validity reward, an execution success reward, and a diversity bonus:
\vspace{-0.2cm}
\begin{equation}
    R_{\text{tool}}(\tau) = \mathbb{1}_{\text{syntax}}(\tau) + \mathbb{1}_{\text{success}}(\tau) + R_{\text{div}}(\tau)
\end{equation}
where $\mathbb{1}_{\text{syntax}}$ and $\mathbb{1}_{\text{success}}$ are binary indicators ensuring the tool call follows the correct format and returns a successful response (non-error), respectively. To prevent mode collapse (e.g., relying solely on OCR) and to penalize infinite loops of tool invocations, the diversity reward $R_{\text{div}}$ incentivizes the use of unique tools subject to two constraints: a per-tool saturation cap $\kappa$ and a global diversity cap $\eta$:
\vspace{-0.2cm}
\begin{equation}
\small
    R_{\text{div}}(\tau) = \min \left( \eta, \sum_{T_k \in {T}} \beta \cdot \min(\text{count}(T_k, \tau), \kappa) \right)
\end{equation}
Here, $\beta$ is the per-tool reward and $\eta$ limits the total diversity reward to prevent reward hacking via excessive tool calls.

\paragraph{Terminal Delimitation ($R_{\text{term}}$).} A structural anchor ensuring the reasoning process converges to a definitive answer token (<answer>), preventing infinite reasoning loops:
\vspace{-0.2cm}
\begin{equation}
    R_{\text{term}}(\tau) = C2 \cdot \mathbb{1}(\langle \text{answer} \rangle \in \tau)
\end{equation}
where $C2 \in \mathbb{Z}$ represents the reward magnitude.

Finally, to achieve stabilized advantage updates, we normalize and scale the total reward function. As a result, the final reward function is:
\vspace{-0.2cm}
\begin{equation}
    R_{\text{Total}}(\tau) =  S \times \frac{R_{\text{total}}(\tau)}{ N_{\text{norm}}}
\end{equation}
where $N_{\text{norm}} = \lvert \max(R_{\text{total}}) \rvert$ is the normalization factor, which is the maximum possible reward allowed, and $S$ is the scaling factor. The nomenclature of all notations is detailed in Appendix \ref{app:nomenclature}.

\section{Evaluation Measures}
\subsection{Task Accuracy}
We evaluate and report the specific task accuracies within the context of MCQ-style Q/A setup.

\begin{table*}[t]
\centering
\small
\resizebox{\textwidth}{!}{
\begin{tabular}{lccccccc}
\toprule
\textbf{Model} & \textbf{AI2D} & \textbf{TQA} & \textbf{OK-VQA} & \textbf{SCIENCE-QA} & \textbf{Avg. (In Dist.)} & \textbf{MMMU-Pro} \\
\midrule

\rowcolor{basecolor}
\multicolumn{7}{c}{\textbf{Closed-Source Models}} \\
GPT-4o & 76.5 & 77.0 & 88.5 & 86.0 & 82.0 & 61.8 \\
GPT-4o Mini & 64.0 & 73.5 & 78.5 & 83.5 & 74.9 & 51.9 \\

\midrule
\rowcolor{ablatecolor}
\multicolumn{7}{c}{\textbf{Base Models (Non-Agentic)}} \\
Qwen2.5-VL [3B] & 42.2 & 42.2 & 44.5 & 42.0 & 42.7 & 32.7 \\
Gemma-3-IT [4B] & 68.0 & 73.0 & 74.0 & 73.0 & 72.0 & 42.0 \\
Phi-4-Multimodal [6B] & 50.0 & 52.0 & 65.0 & 73.5 & 60.1 & 41.2 \\
Qwen2.5-VL [7B] & 63.8 & 74.6 & 71.5 & 73.5 & 70.9 & 40.5 \\

\midrule
\rowcolor{toolcolor}
\multicolumn{7}{c}{\textbf{Baseline (Agentic)}} \\
VERL: Qwen2.5-VL [3B] & 56.5 & 56.0 & 70.5 & 58.0 & 60.3 & 33.8 \\
Custom: Gemma-3-IT [4B] & 59.5 & 68.0 & 74.0 & 72.0 & 68.4 & 30.2 \\
Custom: Phi-4-Multimodal [6B] & 19.5 & 20.0 & 19.0 & 26.5 & 20.1 & 16.9 \\
HF SmolAgent: Qwen2.5-VL [7B] & 66.0 & 72.5 & 76.0 & 79.0 & 73.4 & 36.9 \\
VERL: Qwen2.5-VL [7B] $\uparrow$ & 67.5 & 73.3 & 74.6 & 78.3 & 73.4 & 44.3 \\

\midrule
\rowcolor{spectraheading}
\multicolumn{7}{c}{\textbf{\spectra\ (Ours)}} \\
\rowcolor{spectracolor}
VERL: Qwen2.5-VL [3B] & \textbf{60.5} & \textbf{60.0} & \textbf{66.5} & \textbf{68.5} & \textbf{63.9} & \textbf{36.3} \\
\rowcolor{spectracolor}
VERL: Qwen2.5-VL [7B] & \textbf{71.1} & \textbf{77.5} & \textbf{79.6} & \textbf{83.1} & \textbf{77.8} & \textbf{46.7} \\

\bottomrule
\end{tabular}
}
\caption{Benchmark comparison of \textbf{\spectra} over various baselines (Agentic and Non-Agentic Zero-shot), including out-of-distribution MMMU-Pro for both the variants: Qwen2.5-VL (3B and 7B). Where, $\uparrow$ denotes closest baseline.}
\label{tab:model_comparison}
\end{table*}

\subsection{Tool Instrumental Utility (TIU)}
\label{sec:tiu_metric}
To rigorously evaluate the efficacy of tool use in agentic models under self-supervision, we define an evaluation set $\mathcal{D}_{\text{eval}}$ and utilize the library of available tool primitives ${T}$ (as defined in Section \ref{sec:methodology}). We propose \textit{Tool Instrumental Utility} (TIU) as a composite metric to evaluate the overall effectiveness of tool use in the absence of human preferences. TIU integrates reliability, alignment, and selectivity into a normalized scalar $[0, 1]$. It is defined as:
\vspace{-0.2cm}
\begin{equation}
\small
    \text{TIU} = \text{TER} \times \underbrace{\left( \frac{1 + \text{TTAC}}{2} \right)}_{\text{Normalized Relevance}} \times \underbrace{\tanh(\text{TSS})}_{\text{Bounded Intentionality}}
\end{equation}
The term $\frac{1 + \text{TTAC}}{2}$ normalizes the alignment coefficient to $[0, 1]$ for the metric Task-Tool Alignment Coefficient (TTAC), penalizing detrimental tool use. The hyperbolic tangent $\tanh(\text{TSS})$ bounds the infinite range of the KL-divergence based Tool Selectivity Score (TSS), rewarding selectivity with diminishing returns. A TIU score of 1.0 represents an agent that is perfectly reliable, strategically selective, and uses tools that guarantee task success. We describe the components of this metric below:

\paragraph{Tool Execution Reliability (TER).} TER measures the mechanical robustness of the agent, specifically the rate of syntactically valid and error-free tool invocations. Let $N$ be the total number of tool calls attempted across $\mathcal{D}_{\text{eval}}$. It is defined as:
\vspace{-0.2em}
\begin{equation}
\small
    \text{TER} = \frac{1}{N} \sum_{i=1}^{N} \mathbb{1}(o_i = \text{success})
    \vspace{-0.2cm}
\end{equation}
where $\mathbb{1}$ is the indicator function and $o_i$ represents the execution outcome of the $i$-th tool call.

\paragraph{Task-Tool Alignment Coefficient (TTAC).} To assess instrumental relevance, we quantify the relationship between tool usage and task success. TTAC is calculated as the average point-biserial correlation between the binary usage vector $\mathbf{u}_T$ (indicating if tool $T$ was used) and the task success vector $\mathbf{y}_{\text{succ}}$ (derived from $R_{\text{corr}}$):
\vspace{-0.2em}
\begin{equation}
\small
    \text{TTAC} = \frac{1}{|{T}|} \sum_{T_k \in {T}} r(\mathbf{u}_{T_k}, \mathbf{y}_{\text{succ}})
\end{equation}
Here, positive values of Pearson correlation coefficient $r \in [-1, 1]$ indicate that tool $T$ usage correlates with successful task resolution.

\paragraph{Tool Selectivity Score (TSS).} TSS measures the agent's intentionality by quantifying the deviation of its empirical tool usage distribution $P$ from a random uniform distribution $U$. We utilize the Kullback-Leibler (KL) divergence:
\vspace{-0.2em}
\begin{equation}
\small
    \text{TSS} = D_{KL}(P || U) = \sum_{T_k \in {T}} P(T_k) \log\left(\frac{P(T_k)}{U(T_k)}\right)
\end{equation}
Higher values indicate strong preferences for strategic selection of tools within the action space $\mathcal{A}$, while values approaching 0 indicate random guessing. Please refer to Appendix \ref{app:tiu} for more details.

\section{Experimental Results}
\paragraph{Experimental Setting.} We implement \spectra by fine-tuning Qwen2.5-VL (3B and 7B) \cite{bai2025qwen25vltechnicalreport} using the VERL framework \cite{sheng2024hybridflow} with the vLLM engine \cite{kwon2023efficient}. We apply LoRA adapters \cite{hu2021loralowrankadaptationlarge} to the language model while freezing the vision encoder. 
We evaluate performance against strong baselines on both in-distribution and out-of-distribution (MMMU-Pro) test data to demonstrate scalability. Prompts, hyperparameters, and computational efficiency details are provided in Appendices \ref{app:prompts}, \ref{app:hyperparameter}, and \ref{app:compeff}, respectively.

\paragraph{Baselines.} Our evaluation utilizes Phi-4-Multimodal-Instruct [6B] \cite{microsoft2025phi4minitechnicalreportcompact}, Gemma-3-IT [4B] \cite{gemmateam2025gemma3technicalreport}, and a closed-source GPT-4o \cite{openai2024gpt4ocard}. We benchmark \spectra\ against both standard (non-agentic) base models and strong agentic baselines including smolagents \cite{smolagents}. For rigorous qualitative assessment, we compare \spectra with the closest VERL-based agentic baseline using Qwen2.5-VL [7B].
\begin{table}[!t]
\centering
\resizebox{\linewidth}{!}{
\begin{tabular}{llcccc}
\toprule
\textbf{Model} & \textbf{Dataset} 
&\makecell{\textbf{TER (\%)}} & \makecell{\textbf{ TTAC}} & \makecell{\textbf{TSS}} & \makecell{\textbf{TIU (\%)}} \\ 
\midrule
\multirow{5}{*}{\shortstack[l]{Baseline\\Agent}} 
 & AI2D & 77.08 & -0.051 & 1.32 & 31.77 \\
 & TQA & 78.92 & 0.078 &  2.96 & 42.34\\
 & OK-VQA & 80.13 & 0.009 & 2.78 & 40.12 \\
 & SCIENCE-QA & 73.08 & -0.049 & 1.14 & 28.29 \\
 \rowcolor{toolcolor}
 & \textbf{Mean} & \textbf{77.30} & \textbf{-0.003} & \textbf{2.05} & \textbf{35.63} \\ 
\midrule
\multirow{5}{*}{\textbf{\spectra}} 
 & AI2D & 85.71 & -0.025 & 3.08 & 41.59 \\
 & TQA & 90.07 & 0.081 & 3.04 & 48.47 \\
 & OK-VQA & 85.16 & -0.100 & 2.76 & 37.99 \\
 & SCIENCE-QA & 93.84 & 0.082 & 3.05 & 50.58 \\
 \rowcolor{spectracolor}
 & \textbf{Mean} & \textbf{88.69} & \textbf{0.009} & \textbf{2.98} & \textbf{44.66} \\ 
\bottomrule
\end{tabular}
}
\caption{Tool efficiency enhancement in \spectra\ 7B over baseline agent across datasets. (TER: Tool Execution Reliability, TTAC: Tool-Task Alignment Coefficient, TSS: Tool Selectivity Score, and TIU: Tool Instrumental Utility; a composite metric)}
\label{tab:tiu_results}
\end{table}

\begin{figure}[!t]
 \centering
\includegraphics[width=0.99\columnwidth]{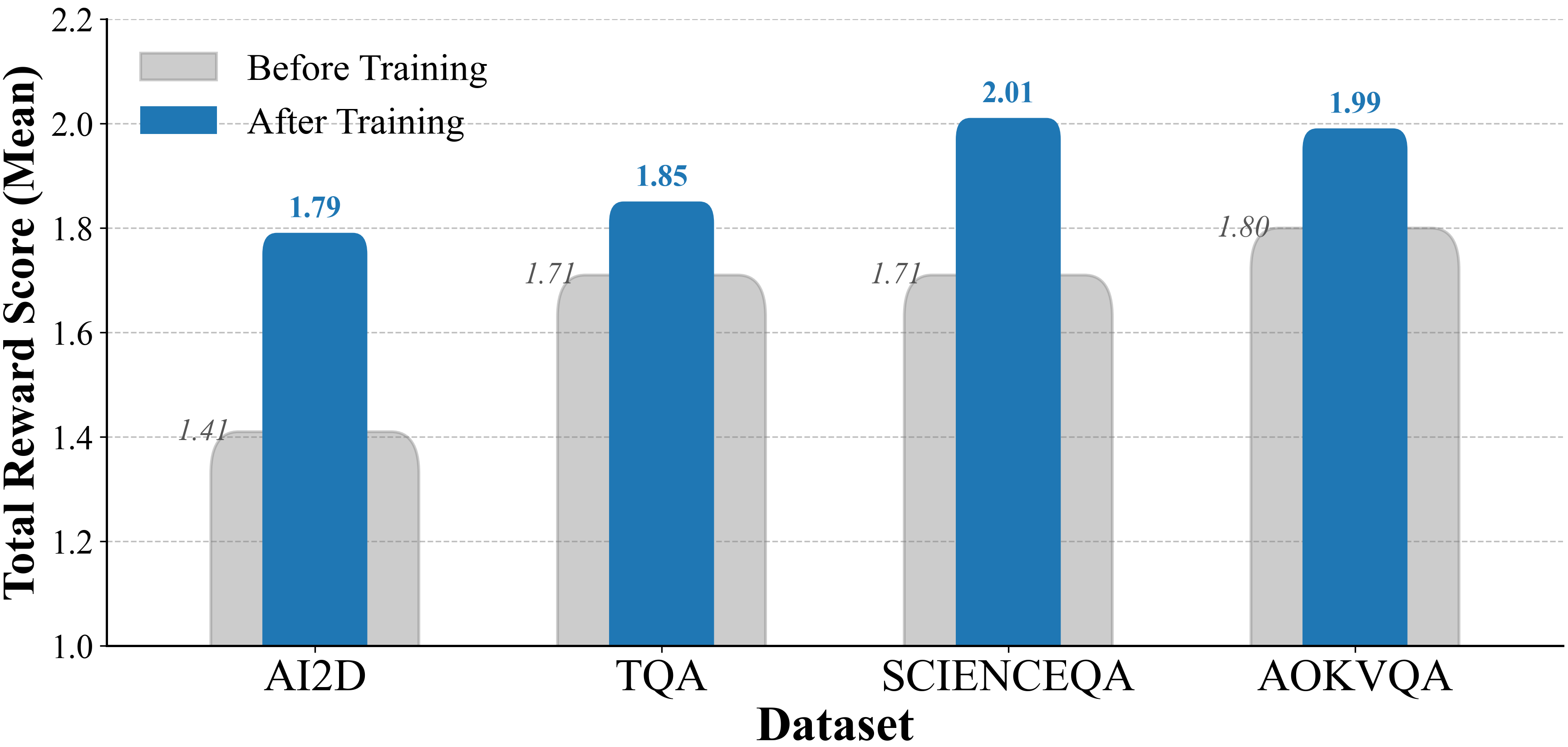}
  \caption{Average total reward progression across datasets after applying \spectra's (7B) optimization.} 
  \label{fig:reward_update}
\end{figure}

\subsection{Benchmark Comparison}
As shown in Table \ref{tab:model_comparison}, \spectra\ 3B and 7B consistently outperform the closest strong VERL-based agentic baseline across all benchmarks, demonstrating strong generalization on both in-distribution and the out-of-distribution (MMMU-Pro) datasets. Notably, \spectra\ achieves an increment of 3 to 5 percentage points across benchmarks and a substantial gain on reasoning-intensive tasks, surpassing the baseline by margins of $4.8$ and $5.0$ percentage points on SCIENCE-QA and OK-VQA, respectively for 7B variant. To validate the robustness of these findings, we report the mean average over three independent runs for \spectra\ 7B and the baseline; an independent two-tailed T-test, average across datasets, $p$-value of $0.0019$ confirms the statistical significance of the finding.
\begin{table}[!t]
    \centering
    \small
    \resizebox{\linewidth}{!}{%
        \begin{tabular}{lccccc}
            \toprule
            \textbf{Setting} & \textbf{AI2D} & \textbf{TQA} & \textbf{OKVQA} & \textbf{ScienceQA}& \textbf{Avg.}  \\
            \midrule
            $R_{\text{total}}$ & 71.17 & 77.5 & 79.67 & 83.17 & 77.8 \\
            \midrule
            w/o $R_{\text{corr}}$ & \cellcolor{red!22}68.5 & \cellcolor{gray!15}78.5 & \cellcolor{gray!15}80.5 & \cellcolor{red!41}77.5 &\cellcolor{red!14}76.2\\
            
            w/o $R_{\text{struct}}$ & \cellcolor{red!38}66.0 & 77.5 & \cellcolor{gray!15}82.5 & \cellcolor{red!45}77.0 &\cellcolor{red!18}75.7\\
            
            w/o $R_{\text{tool}}$ & \cellcolor{gray!15}74.5 & \cellcolor{red!28}74.0 & \cellcolor{red!6}79.5 & \cellcolor{red!38}78.0 &\cellcolor{red!14}76.5 \\
            
            w/o $R_{\text{term}}$ & \cellcolor{gray!15}72.0 & \cellcolor{red!18}75.5 & \cellcolor{red!19}77.5 & \cellcolor{red!38}78.0 &\cellcolor{red!18}75.7\\
            \bottomrule
        \end{tabular}%
    }
    \caption{Leave-one-out ablation study of reward components in \spectra\ 7B. Performance changes are visualized via background color: gray indicates improvement, while red indicates a drop compared to total reward.}
    \label{tab:ablation_results}
    \vspace{-0.1cm}
\end{table}
\begin{figure}[!t]
 \centering
 \setlength{\fboxsep}{0pt}  
 \setlength{\fboxrule}{0.1pt} %
\fbox{\includegraphics[width=0.99\columnwidth]{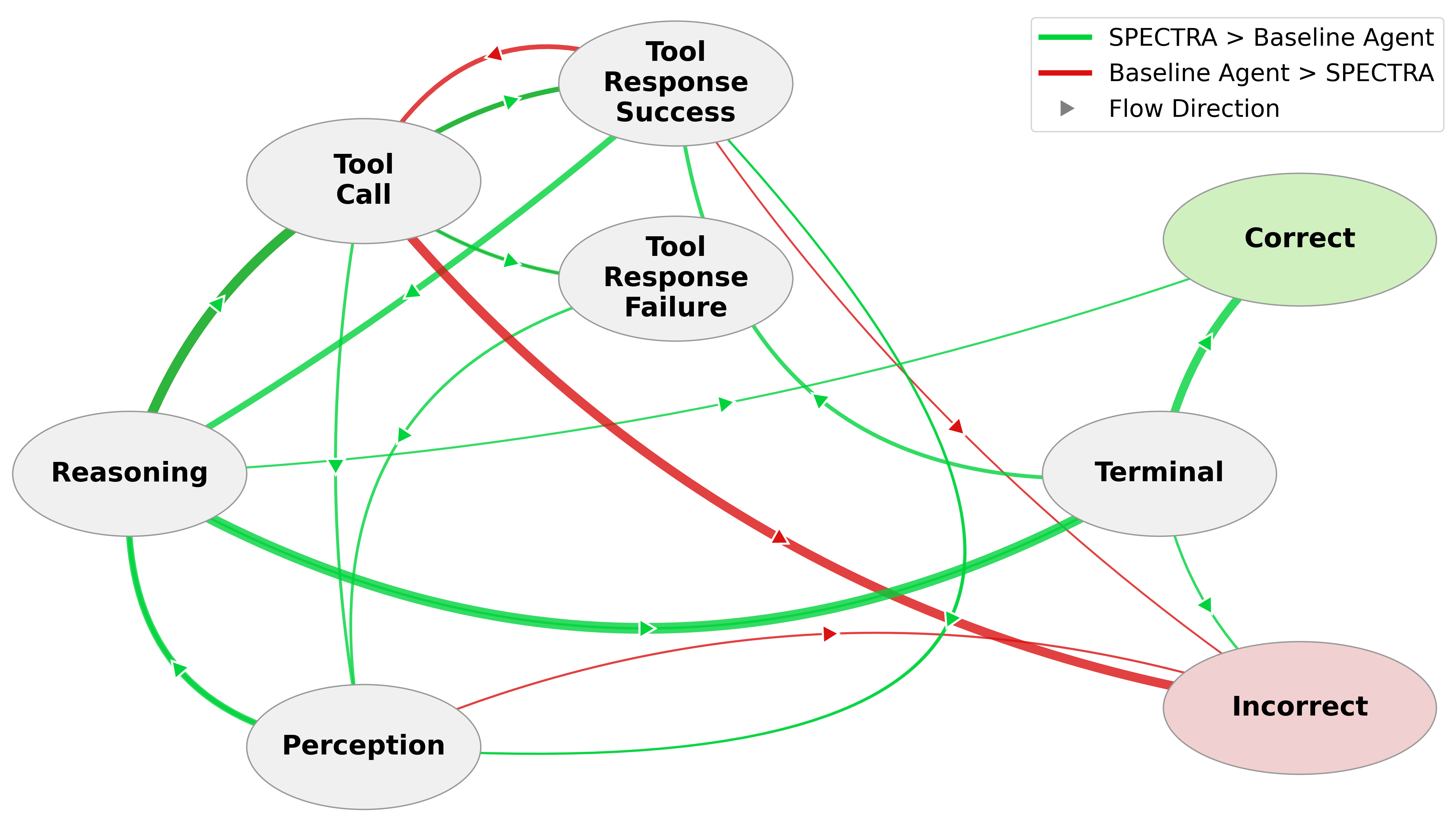}}
  \caption{Differential State Transition Analysis: Baseline agent vs. \spectra\ 7B (in-distribution test set). The impactful transitions shifts are highlighted by the magnitude of color change. \spectra\ incentivizing green paths and down-scaling red paths w.r.t. baseline agent.} 
  \label{fig:state_transition}
  \vspace{-0.4cm}
\end{figure}

\begin{figure*}[t!]
    \centering
    \small
    \begin{mainbox}{Comparative Analysis: Robustness to Execution Errors}
        \vspace{-0.2cm}
        \begin{minipage}{0.72\textwidth}
            \footnotesize
            \textbf{Question:} In the food web, what role does grass play? \\
            \textbf{Options:} A) Producer/Energy Source \quad B) Decomposer \quad C) Predator \quad D) Consumer
        \end{minipage}%
        \hfill
        \begin{minipage}{0.25\textwidth}
            \centering
        \includegraphics[height=1.8cm, keepaspectratio]{test_123_ai2d_test_image_56.png}  
            \vspace{-0.3cm}
        \end{minipage}        

        \begin{minipage}[t]{0.48\textwidth}
            \begin{subbox}{failbg}{failframe}{Baseline Agent}
                \footnotesize
                \textbf{Reasoning:} Identifies grass as producer.
                \vspace{0.05cm}
                
                \toolcall{>> Tool Call: None()} \\
                \toolcall{<< Err: Exec tool 'None'}
                
                \vspace{0.05cm}
                \textbf{Crash:} The agent enters a loop calling `None` and terminates abruptly.
                \vspace{0.1cm}
                
                \textbf{Final Output:} \\
                \texttt{assistant\textbackslash n <think} ... \textit{[End of context]}
                
                \vspace{0.2cm}
                \textbf{Prediction:} \textcolor{red}{\textbf{None (Failed)}}
            \end{subbox}
        \end{minipage}%
        \hfill
        \begin{minipage}[t]{0.48\textwidth}
            \begin{subbox}{passbg}{passframe}{\spectra\ (Ours)}
                \footnotesize
                \textbf{Reasoning:} Identifies grass as producer.
                \vspace{0.05cm}
                
                \toolcall{>> Tool Call:\{"captioning\_tool"...\}} \\
                \hallucination{(Undefined <Captioning\_Tool> Tag Hallucination)}
                \hallucination{addCriterion: The image is...} \\
                \hallucination{[Repetitive Loop x3]}
                
                \vspace{0.05cm}
                \textbf{Recovery:} Despite the hallucinated tool and syntax errors, it recovers to output the answer.
                \vspace{0.05cm}
                
                \textbf{Prediction:} \boxed{\text{A}} \textcolor{green!50!black}{\textbf{\checkmark}}
            \end{subbox}
        \end{minipage}
    \vspace{-0.2cm}
    \end{mainbox}
    \vspace{-0.3cm}
    \caption{Robustness comparison. The \textbf{Baseline Agent} (left) crashes after a tool error, ending abruptly without an answer. \textbf{\spectra} (right) hallucinates an undefined tool (highlighted in red) and enters a generation loop, yet successfully recovers to provide the correct answer (A).}
    \label{fig:error_ana}
    \vspace{-0.5cm}
\end{figure*}

\subsection{Analysis of Tool Utility and Efficiency}
Quantitative analysis presented in Table \ref{tab:tiu_results} reveals that \spectra\ significantly enhances agentic proficiency, achieving a $25.3\%$ relative improvement in mean TIU compared to the baseline ($35.63\% \rightarrow 44.66\%$). This gain is predominantly driven by a robust $11.4\%$ increase in TER, confirming that our topological constraints effectively minimize syntax errors and invalid calls, particularly in ScienceQA ($93.84\%$). Furthermore, the elevated TSS of $2.99$ (vs.~$2.06$) indicates that the policy successfully avoids mode collapse, converging on a focused subset of high-utility tools rather than random guessing. While TTAC remains a challenging objective under cold-start conditions, \spectra\ notably shifts the mean correlation from negative to positive ($0.010$), suggesting that self-discovered policies effectively ground visual reasoning in actionable evidence without supervised preferences.

\subsection{Agentic Trajectory Analysis}
We perform a differential state transition analysis in multi-turn rollout to evaluate \spectra\ against the baseline agent, highlighting shifts in error propagation and resolution efficiency. Results presented in Figure \ref{fig:state_transition} demonstrate that \spectra\ significantly optimizes the trajectory toward successful outcomes, evidenced by a marked increase in \texttt{Reasoning} $\to$ \texttt{Terminal} transitions ($+48$) and \texttt{Reasoning} $\to$ \texttt{Tool\_Call} accuracy ($+44$). Notably, the model consolidates the iterative loop \texttt{Reasoning} $\to$ \texttt{Tool\_Call} $\to$ \texttt{Tool\_Response\_Success} $\to$ \texttt{Perception} $\to$ \texttt{Reasoning} as the primary success path, verifying its enhanced stability in multi-step execution. Crucially, the model actively curtails pathological behaviors observed in the baseline, specifically suppressing the recursive, error-prone \texttt{Tool\_Call} $\to$ \texttt{Tool\_Call} loops ($-103$) and reducing direct \texttt{Tool\_Call} $\to$ \texttt{Incorrect} failures ($-41$). This structural shift indicates that \spectra\ dampens cyclic dependencies in the incorrect category while incentivizing linear, reasoning-driven paths to correct terminal states. The detailed quantitative results are provided in Appendix \ref{app:agentic_trajectory}.

\subsection{Reward Investigation}
\paragraph{Leave-One-Out Reward Ablation.} The leave-one-out ablation study (Table \ref{tab:ablation_results}) highlights the effectiveness of \spectra’s reward formulation, especially on ScienceQA, where removing any component causes a performance drop exceeding $5\%$. The impact of $R_{\text{tool}}$ varies across datasets, reflecting differing levels of tool reliance in tasks such as AI2D and TQA. The structural integrity reward ($R_{\text{struct}}$) remains critical for preventing reasoning collapse in complex visual question answering. Overall, robust generalization requires the full multi-objective framework despite domain-sensitive sub-objectives.

\paragraph{Reward Convergence Findings.} In Figure \ref{fig:reward_update}, \spectra\ demonstrates substantial and consistent performance gains across all evaluated datasets, with the most notable improvement in AI2D, where the average reward increases from $1.41$ to $1.79$ ($+27\%$). This uniform uplift validates that the newly formulated multi-objective agentic reward effectively stabilizes the optimization process, allowing the model to internalize complex multimodal reasoning constraints without the high variance typical of traditional reinforcement learning 
across diverse tasks ranging from diagram interpretation (AI2D) to knowledge-intensive QA (TQA).


\subsection{Qualitative Error Analysis}
\spectra\ significantly improves over the baseline by optimizing tool selection, enforcing structured multi-turn rollouts, and ensuring convergence to terminal conditions, effectively preventing infinite loops or abrupt endings. However, as illustrated in Figure \ref{fig:error_ana}, the formulation is not without limitations; it occasionally exhibits intermediate instability, such as hallucinating tool tags or generating repetitive text, before arriving at the correct solution. Appendix \ref{app:case_studies} provides comprehensive qualitative examples of error recoveries and failures, while Appendix \ref{app:tool_ter} contains the tool-level analysis.

\section{Conclusion}
We propose \spectra, a novel supervision-free framework that uses Coldstart RL to optimize SVLM tool use and visual reasoning. By introducing multi-turn topological constraints with a multi-objective reward, \spectra\ improves grounded reasoning and mitigates perceptual limitations. We further introduce Tool Instrumental Utility, a metric for quantifying tool efficiency without supervised preferences. Experiments show that environment-driven signals combined with structured constraints significantly enhance SVLM agentic performance, pointing toward autonomous, perceptually aware agents without costly human supervision.

\section{Limitations}
While \spectra\ demonstrates significant advancements in visual-intensive tasks, there are two primary limitations to our current approach. First, the model is explicitly designed with a focus on vision-specific tools. Consequently, it lacks access to broader utilities such as code execution environments or search engines, which can be critical for solving complex, multi-modal vision tasks requiring external knowledge or precise calculation. Future iterations of \spectra\ could integrate these general-purpose tools to expand its applicability beyond strictly vision-focused analysis.

Second, despite substantial improvements over strong baselines, \spectra\ occasionally exhibits hallucinations within intermediate reasoning steps, even when the final prediction is correct. This phenomenon indicates that while the outcome is robust, the interpretability and factual grounding of the reasoning chain can be further refined. Future work will explore the integration of specific hallucination-removal objectives and consistency constraints to mitigate these errors and ensure high fidelity across the entire reasoning process.

\section{Ethical Considerations}
This study relies exclusively on publicly available datasets for the training and evaluation of \spectra. These resources were utilized solely for scientific research and academic purposes. We explicitly state that we do not claim ownership or licensing rights over any of the datasets employed in this work. All data remains subject to the original licenses, terms of use, and copyright protections established by the respective source providers. We have made every effort to ensure our usage complies with these original distribution terms. In accordance with AI usage guidelines, generative AI tools were used solely for the purposes of paraphrasing, grammar correction, and spell checking.

\section*{Acknowledgments}
Tanmoy Chakraborty acknowledges the support of Microsoft Research Grant, Azure AI Credits Grant from Microsoft's Accelerating Foundation Models Research (AFMR) initiative, Google GCP Grant, and Rajiv Khemani Young Faculty
Chair Professorship in Artificial Intelligence.

\bibliography{camera_ready_paper}

\appendix
\section{Benchmark Details}
\label{app:benchmarks}
This section details the public datasets utilized for the training and in- and out-of-distribution evaluation of \spectra. A sample from each dataset is illustrated in Figure \ref{fig:dataset_illustration}. Here is the details of datasets involved in \spectra's training and in-distribution evaluations:
\begin{figure*}[!t]
    \centering
    \makebox[\textwidth][c]{    \includegraphics[width=1.40\textwidth]{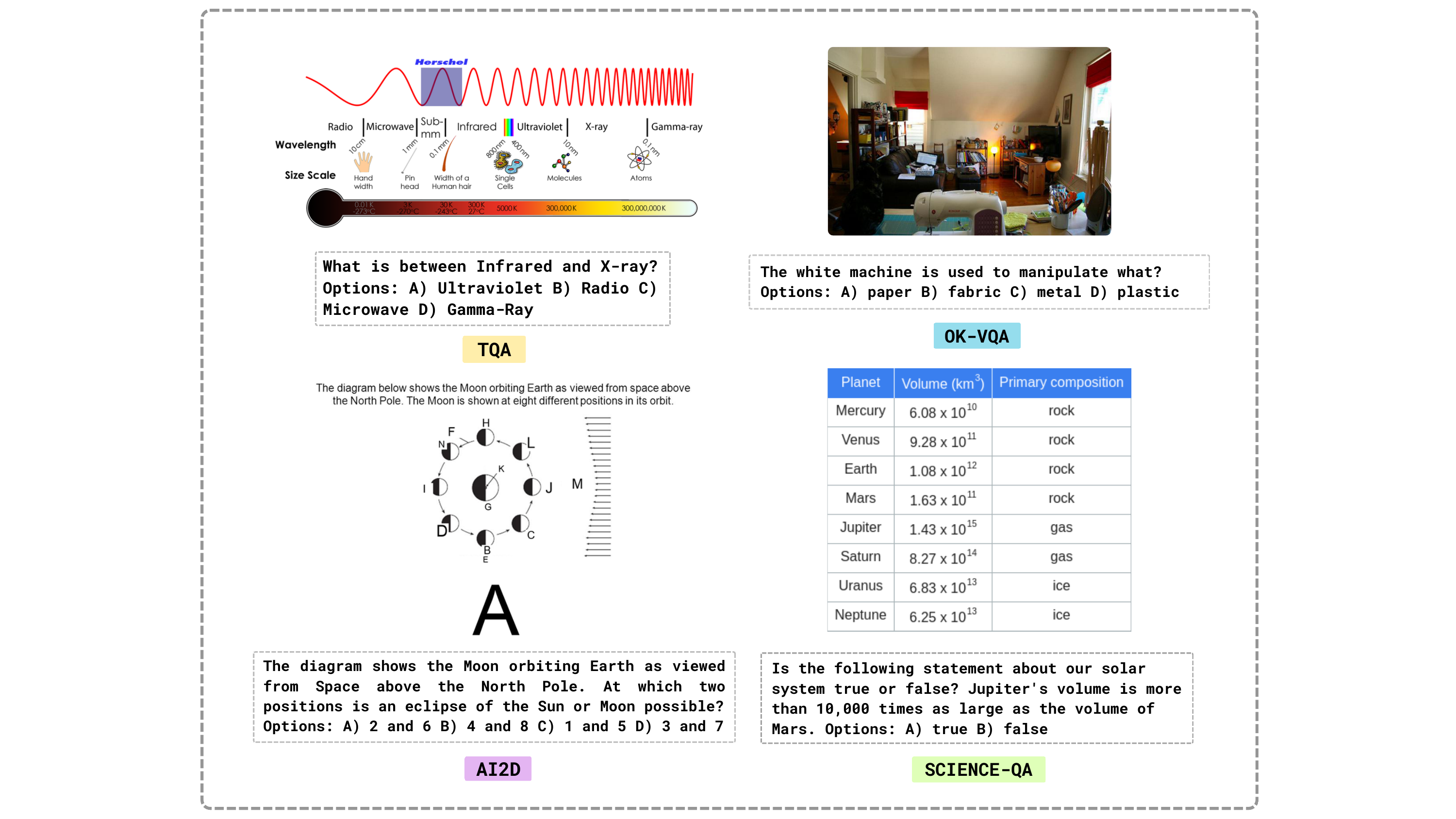}}
    \caption{A sample from each of the four datasets (AI2D, TQA, OK-VQA, and Science QA) used in \spectra's training and in-distribution evaluation.}
\label{fig:dataset_illustration}
\end{figure*}
\begin{itemize}
    \item \textbf{Allen Institute for Artificial Intelligence Diagrams (AI2D)}: AI2D \cite{kembhavi2016diagramworthdozenimages} dataset contains grade school level textbook style scientific diagrams paired with multiple choice questions that evaluate diagram understanding and visual reasoning. Contains questions from physics, biology and Earth and environmental science.
    \item \textbf{Textbook Question Answering (TQA)}: TQA \cite{kim2019textbookquestionansweringmultimodal} Textbook Question Answering (TQA) dataset is a multimodal benchmark constructed from middle school science textbooks that combines text passages, diagrams, and multiple-choice questions. Questions are from physics, biology and geography.
    \item \textbf{Outside Knowledge Visual Question Answering (OK-VQA)}: OK-VQA \cite{marino2019okvqavisualquestionanswering} is a visual question answering dataset where answering questions about images requires external world knowledge beyond what is visible in the image. The questions cover broad domains such as commonsense, culture, geography, history, and objects, encouraging models to combine vision with knowledge-based reasoning.
    \item \textbf{Science Question Answering (ScienceQA)}: ScienceQA \cite{lu2022learnexplainmultimodalreasoning} is a large multimodal question answering dataset built from grade-school science curricula, containing questions with images, diagrams, text explanations, and multiple-choice answers. The dataset spans subjects such as natural science, social science, and language arts, and emphasizes reasoning, explanation, and multimodal understanding.
    \end{itemize}
    
\paragraph{Out-Of-Distribution Evaluation.} To assess \spectra's out-of-distribution robustness, we evaluate it against Massive Multi-discipline Multimodal Understanding and Reasoning Pro version (MMMU-Pro) \cite{yue2025mmmuprorobustmultidisciplinemultimodal}, a contemporary multimodal benchmark requiring complex multi-discipline reasoning. The dataset is an enhanced iteration of the original MMMU dataset spanning 30 college-level subjects. This benchmark is designed to filter out text-only biases and prioritize tasks that require deep integration of visual and textual information, reflecting real-world academic challenges. For our evaluation, we focused on the standard setting (four-option multiple choice) and specifically isolated single-image questions to maintain consistency with our training environment. This refined subset results in a final testing dataset of 1,592 (out of 1730) data points, ensuring a focused and robust assessment of the model’s multimodal capabilities.

\section{Tools}
To enable robust multimodal processing, \spectra\ integrates a specialized set of tools targeting distinct visual capabilities. High-level semantic context is derived using the Image Captioning Tool for scene summarization, while fine-grained entity localization is achieved via the Object Detection Tool. Textual information is extracted by the OCR Tool, and complex spatial reasoning is handled by the Visual Perception Tool, powered by a VLM, ensuring comprehensive scene interpretation and detail analysis. Respective implementation details for each tool is presented at Table \ref{tab:tooldesc}.
\begin{table*}[h]
    \centering
    \small
    \resizebox{\textwidth}{!}{
    \begin{tabular}{l l p{8cm}}
        \toprule
        \textbf{Tool Name} & \textbf{Method} & \textbf{Description} \\
        \midrule
        Image\_Captioning\_Tool & BLIP2 \cite{10.5555/3618408.3619222} & Generates a natural language summary describing the visual content of an image. \\
        & & \textit{Response}: plain-text\\
        Object\_Detection\_Tool & DETR \cite{carion2020endtoendobjectdetectiontransformers} & Identifies and localizes specific objects within an image using bounding boxes. \\
        & & \textit{Response}:box: [x1, y1, x2, y2], label: string, score: float \dots \\
        OCR\_Tool & Tesseract \cite{10.5555/1304596.1304846} & Extracts and digitizes printed or handwritten text characters from visual inputs. \\
        & & \textit{Response}: plain-text\\
        Visual\_Perception\_Tool & Qwen2.5-VL[7B] instruct \cite{bai2025qwen25vltechnicalreport} & Analyzes images to generate spatial relationships or details. \\
        & & \textit{Response}: plain-text\\
        \bottomrule
    \end{tabular}
    }
    \caption{Description of Tools.}
     \label{tab:tooldesc}
\end{table*}
\section{Methodology And Evaluation Measures Cont.}
In this section we provide additional details that need derivation and disclosure for reproducibility.

\subsection{Hierarchical Soft Structural Integrity Reward Details}
\label{app:reward_struct}
The hierarchical soft structural integrity reward term is given by:
\begin{equation}
R_{\text{struct}}(\tau) = \alpha \cdot \gamma^{\phi(\tau)}    
\end{equation}
where $\alpha=2.0$, $\gamma=0.75$, and $\phi(\tau)$ map partial structures (e.g., missing perception) to higher integers.

Our motivation is to maximize rewards for the optimal multi-turn rollout structure without penalizing other valid paths that reach the correct solution. This approach ensures the model captures learning signals from various successful topological constraints, encouraging it to discern when tool usage is unnecessary.

We define three rollout structure templates as follows:

\paragraph{Template: Optimal ($Z_1$)}
\begin{center}    
\begin{tikzpicture}
\node[
  draw,
  dashed,
  rectangle,
  inner sep=6pt,
  anchor=west,
  align=left
] {
\small
\texttt{<think\_reasoning> ... </think\_reasoning>}\\
\small
\texttt{\ \ \ \ <tool\_call> ... </tool\_call>}\\
\small
\texttt{\ \ \ \ <tool\_response> ... </tool\_response>}\\
\small
\texttt{<think\_perception> ... </think\_perception>}\\
\small
\texttt{<think\_reasoning> ... </think\_reasoning>}\\
\small
\texttt{.....}\\
\small
\texttt{<answer> ... </answer>}
};
\end{tikzpicture}
\end{center}

\paragraph{Template: Valid ($Z_2$)}
\begin{center}    
\begin{tikzpicture}
\node[
  draw,
  dashed,
  rectangle,
  inner sep=6pt,
  anchor=west,
  align=left
] {
\small
\texttt{<think\_reasoning>  ... </think\_reasoning>}\\
\small
\texttt{<think\_perception> ... </think\_perception>}\\
\small
\texttt{<think\_reasoning>  ... </think\_reasoning>}\\
\small
\texttt{<answer> ... </answer>}
};
\end{tikzpicture}
\end{center}

\paragraph{Template: Alternative ($Z_3$)}
\begin{center}    
\begin{tikzpicture}
\node[
  draw,
  dashed,
  rectangle,
  inner sep=6pt,
  anchor=west,
  align=left
] {
\small
\texttt{<think\_reasoning>  ... </think\_reasoning>}\\
\small
\texttt{<answer> ... </answer>}
};
\end{tikzpicture}
\end{center}

We define the reward signals for the three template structures based on the parameterization shown in Table \ref{tab:reward_params}. The reward contribution is zero in case model follows any deviations other than the defined ones. 

\begin{table}[h]
\small
\centering
\begin{tabular}{lcccc}
\toprule
\textbf{Template} & $\alpha$ & $\gamma$ & $\phi(\tau)$ & \makecell{\textbf{$R_{struct}$}} \\ 
\midrule
$Z_1$ (Optimal)     & 2.0      & 0.00     & 0            & 2.000                           \\
$Z_2$ (Valid)       & 2.0      & 0.75     & 1            & 1.500                           \\
$Z_3$ (Alternative) & 2.0      & 0.75     & 2            & 1.125                           \\ 
\bottomrule
\end{tabular}
\caption{Reward Signal Parameterization for Rollout Templates}
\label{tab:reward_params}   
\end{table}

\subsection{Tool Selectivity Score (TSS) Details}
\label{app:tiu}
While maximum entropy promotes exploration and randomness \cite{xue2025simpletirendtoendreinforcementlearning}, we minimize entropy to demonstrate tool selectivity. To calculate the TSS, we begin by assuming that the tool selection policy is initially represented as a uniform distribution across a set of $K$ tools. Let ${T} = \{ T_{\text{1}}, T_{\text{2}}, T_{\text{3}}, ... ,T_{\text{K}} \}$ denote the set of |K| independent available tools. The initial distribution of tool selection can be modeled as follows:
\begin{equation}
\small
U(T) = \text{Uniform}({T}) 
\end{equation}
This implies each tool is chosen with equal probability. The probability of each tool being chosen is simply as:
\begin{equation}
\small
U(T_{k}) = \frac{1}{K}    
\end{equation}
Subsequently, we define $P(T_k)$, which represents an empirical tool usage distribution for an agent. This distribution encapsulates the observed likelihood of selecting a given tool $T_k$ based on the agent’s reasoning. It is defined as:
\begin{equation}
\small
P_{\text{}}(T_k) = 
\frac{n_{k}}
{\sum_{j=1}^K n^{\text{}}_j}    
\end{equation}

Here, $n$ represents the total number of times a specific tool is utilized within a given dataset. 
\\
We choose Kullback-Leibler (KL) divergence to model the distribution shift as it effectively represents the agent's specificity in strategically selecting tools, moving away from a random uniform distribution. It is defined as follows:
\begin{equation}
\small
\begin{aligned}
\text{$TSS$} &= D_{KL}(P(T) \, || \, U(T)) = \sum_{T_k \in {T}} 
P(T_k)\,\log\!\left(\frac{P(T_k)}{U(T_k)}\right)
\end{aligned}    
\end{equation}

The range for the TSS is $[0,\infty]$. To constrain the TSS parameter within the interval $[0,1]$, we apply the hyperbolic tangent ($\tanh$) function to the $D_{KL}$ term. The intuition behind the parameter is that the further the distribution deviates from a random uniform distribution, the better. In other words, higher values of tanh$(D_{KL})$, which correspond to greater divergence from randomness, are considered more desirable for an agent to strategically select provided tools. Finally, we also report a mean TSS value over all the datasets under evaluation, defined as:
\begin{equation}
\small
\begin{aligned}
\text{$\bar{TSS}$} &= \frac{1}{|D_{eval}|} \sum_{d \in {D_{eval}}} \sum_{T_k^d \in {T^d}} 
P(T_k^d)\,\log\!\left(\frac{P(T_k^d)}{U(T_k^d)}\right)
\end{aligned}    
\end{equation}

In our experiments, we compared the TSS of \spectra\ with a baseline agent to demonstrate that \spectra\ is better equipped to strategically select tools compared to the baseline agent.

\section{Algorithm}
In Algorithm \ref{alg:cs_arl}, we present the \spectra\ formulation.

\begin{algorithm}[t!]
\small
\caption{\spectrafull}
\label{alg:cs_arl}
\begin{algorithmic}[1]
\REQUIRE Dataset $\mathcal{D}$, SVLM Policy $\pi_\theta$ with frozen $E_\phi$, Toolset ${T}$
\REQUIRE Hyperparameters: Group size $G$, Clip $\epsilon$, Coefficients $\lambda_{1..4}$
\ENSURE Optimized Policy Parameters $\theta^*$

\STATE Initialize LoRA parameters $\Delta\theta$ in $\pi_\theta$
\WHILE{not converged}
    \STATE Sample batch of inputs $\mathcal{B} = \{(I, q)\} \sim \mathcal{D}$
    \FORALL{$(I, q) \in \mathcal{B}$}
        \STATE $\pi_{\theta_{\text{old}}} \leftarrow \pi_\theta$ \COMMENT{Freeze current policy for sampling}
        \STATE \textbf{Step 1: Structured Multi-turn Rollouts}
        \STATE Sample $G$ trajectories $\{\tau_1, \dots, \tau_G\} \sim \pi_{\theta_{\text{old}}}(\cdot | I, q)$
        \STATE \COMMENT{Ensure topology: $\texttt{reason} \to \texttt{tool} \to \texttt{obs} \to \texttt{percep} \to \texttt{reason} \to \texttt{ans}$}
        \STATE \textbf{Step 2: Multi-Objective Reward Calculation}
        \STATE $R_{\text{group}} \leftarrow \emptyset$
        \STATE $N= abs(max(R_{\text{total}}))$ 
        \FOR{$i = 1$ to $G$}
            \STATE Calculate component rewards:
            \STATE \quad $(r_c, r_s, r_t, r_m) \in \tau_i$ 
            \STATE Aggregate, Normalize and Scale: 
            \STATE \quad $R_{\text{total}}(\tau_i) \leftarrow  \lambda_{1} r_c + \lambda_{2}r_s +  \lambda_{3}r_t +  \lambda_{4}r_m$
            
            \STATE \quad $R_{\text{final}}(\tau_i) \leftarrow S * R_{\text{total}}(\tau_i) / N$
            \STATE Append $R_{\text{total}}(\tau_i)$ to $R_{\text{group}}$
        \ENDFOR
        
        \STATE \textbf{Step 3: Trajectory Advantage Estimation}
        \STATE Compute group stats: $\mu, \sigma \leftarrow \text{Mean}(R_{\text{group}}), \text{Std}(R_{\text{group}})$
        \FOR{$i = 1$ to $G$}
            \STATE $\hat{A}_i \leftarrow \frac{R_{\text{total}}(\tau_i) - \mu}{\sigma + \delta}$
        \ENDFOR
    \ENDFOR
    
    \STATE \textbf{Step 4: GRPO Optimization}
    \STATE Compute Loss $\mathcal{J}_{\text{CS-ARL}}(\theta)$
    \STATE Update $\theta \leftarrow \theta + \eta \nabla_\theta \mathcal{L}$
\ENDWHILE
\RETURN $\theta$
\end{algorithmic}
\end{algorithm}
\section{Experimental Results Cont.}
In this section, we present extended results.
\subsection{Tool Utility And Efficiency Cont.}
\label{app:tool_ter}
In this section, we present a qualitative analysis of each tool's performance. Specifically, we evaluate the Tool Execution Reliability (TER) of a tool for \spectra\ in comparison to the baseline agent. The tool-level TER results are presented in Table \ref{tab:tool_level_ter}. The results are averaged over in-distribution test set.

The qualitative analysis demonstrates that \spectra\ generally outperforms the baseline agent, most notably in Visual Perception ($+16.1\%$) and Object Detection ($+9.8\%$). While both agents show nearly identical performance on the Image Captioning Tool, a distinct difference in tool selection strategy is observed regarding OCR. Specifically, \spectra\ does not utilize the OCR Tool, whereas the baseline agent invoked it only twice with a $50\%$ failure rate, suggesting that \spectra's improved performance may stem from a more selective and reliable tool-calling logic.

Additionally, we conducted a human evaluation of tool efficiency, analyzing 50 random samples from the test set. Out of these, tool calls occurred in 40 samples (80\%), with 29 (72.9\%) identified as relevant (considering Tool Efficiency and Tool Selectivity Score metrics) based on human judgment. Of the relevant tool calls, 28 (96.5\%) resulted in correct outcomes, demonstrating a strong correlation between relevant tool calls and favorable results in human verification too.

\begin{table}[!t]
\small
\centering
\begin{tabular}{l|cc}
\toprule
\textbf{Tool} & \multicolumn{2}{c}{\textbf{TER (\%)}} \\
\cmidrule{2-3}
& \textbf{Baseline Agent} & \textbf{\spectra} \\
\midrule
Image\_Captioning\_Tool & 87.2 & 87.8\\
Object\_Detection\_Tool & 65.2 &75.0 \\
OCR\_Tool  & 50.0 & - \\
Visual\_Perception\_Tool &  73.3& 89.4\\
\bottomrule
\end{tabular}
\caption{Tool-level TER statistics for \spectra\ 7B against baseline agent.}
\label{tab:tool_level_ter}   
\end{table}

\subsection{Agentic Trajectory Analysis Cont.}
\label{app:agentic_trajectory}
The detailed quantitative results for agentic trajectory analysis are presented in Table \ref{tab:transition_analysis_full}.

Additionally, we conducted a human-verified qualitative analysis of the <think\_perception> tag on 50 samples, finding favorable outcomes for 40/50 cases (80\%). Among these, 77.5\% followed the <think\_perception> based rollout structure, while 90.3\% of the responses demonstrated high-quality summarization and accurate understanding of image history based on the human judgment. These results confirm the higher likelihood of achieving favorable outcomes when using optimized, task-specific structured rollouts.

\begin{table*}[h]
    \centering
    \small
    \begin{tabular}{l l l r r r}
        \toprule
        \textbf{Source State} & \textbf{Target State} & \textbf{End Outcome} & \textbf{Baseline} & \textbf{\spectra} & \textbf{$\Delta$} \\
        \midrule
        \multicolumn{6}{c}{\textit{\textbf{Top Improvements (Increased Correct Outcomes)}}} \\
        \midrule
        Reasoning & Terminal & \textcolor{green!60!black}{Correct} & 616 & \textbf{664} & \textbf{\textcolor{green!60!black}{+48}} \\
        Reasoning & Tool\_Call & \textcolor{green!60!black}{Correct} & 444 & \textbf{488} & \textcolor{green!60!black}{+44} \\
        Tool\_Call & Tool\_Call & \textcolor{green!60!black}{Correct} & 20 & \textbf{63} & \textcolor{green!60!black}{+43} \\
        Terminal & Correct & \textcolor{green!60!black}{Correct} & 581 & \textbf{618} & \textcolor{green!60!black}{+37} \\
        Perception & Reasoning & \textcolor{green!60!black}{Correct} & 414 & 442 & \textcolor{green!60!black}{+28} \\
        Tool\_Response\_Success & Reasoning & \textcolor{green!60!black}{Correct} & 81 & 109 & \textcolor{green!60!black}{+28} \\
        Tool\_Call & Tool\_Response\_Success & \textcolor{green!60!black}{Correct} & 414 & 436 & \textcolor{green!60!black}{+22} \\
        
        \midrule
        \multicolumn{6}{c}{\textit{\textbf{Top Reductions (Decreased Incorrect Outcomes)}}} \\
        \midrule
        Tool\_Call & Tool\_Call & \textcolor{red}{Incorrect} & 140 & \textbf{37} & \textbf{\textcolor{blue}{-103}} \\
        Reasoning & Tool\_Call & \textcolor{red}{Incorrect} & 167 & 123 & \textcolor{blue}{-44} \\
        Tool\_Call & Incorrect & \textcolor{red}{Incorrect} & 50 & \textbf{9} & \textcolor{blue}{-41} \\
        Tool\_Response\_Success & Tool\_Call & \textcolor{red}{Incorrect} & 33 & 14 & \textcolor{blue}{-19} \\
        Tool\_Call & Tool\_Response\_Success & \textcolor{red}{Incorrect} & 146 & 127 & \textcolor{blue}{-19} \\
        Perception & Incorrect & \textcolor{red}{Incorrect} & 12 & 5 & \textcolor{blue}{-7} \\
        Tool\_Call & Tool\_Response\_Failure & \textcolor{red}{Incorrect} & 9 & 3 & \textcolor{blue}{-6} \\
        
        \midrule
        \multicolumn{6}{c}{\textit{Other Notable Transitions}} \\
        \midrule
        Perception & Reasoning & Incorrect & 81 & 97 & +16 \\
        Terminal & Tool\_Response\_Success & Correct & 36 & 51 & +15 \\
        Tool\_Call & Tool\_Response\_Failure & Correct & 13 & 28 & +15 \\
        Tool\_Call & Perception & Correct & 37 & 48 & +11 \\
        Reasoning & Terminal & Incorrect & 134 & 142 & +8 \\
        \bottomrule
    \end{tabular}
    \caption{Differential State Transition Analysis: Baseline Agent vs. \spectra. The table highlights impactful shifts, sorted by the magnitude of change ($\Delta$). \spectra\ demonstrates a significant reduction in error loops (negative $\Delta$ in Incorrect outcomes) and an increase in successful reasoning paths (positive $\Delta$ in Correct outcomes). The cutoff value for absolute delta magnitude is 5.}
    \label{tab:transition_analysis_full}
\end{table*}
\section{Prompts And Chat Templates}
\label{app:prompts}
\subsection{Prompts}
Figure \ref{fig:zeroshot_prompt} illustrates the system prompt for the zero-shot setting, while Figure \ref{fig:training_prompt} displays the rollout prompt utilized during both training and inference for the agentic setup.

\begin{figure*}[h]
    \centering
    \begin{tcolorbox}[
        colback=gray!5,        
        colframe=black,        
        boxrule=0.8pt,         
        arc=5pt,               
        left=10pt, right=10pt, top=10pt, bottom=10pt, 
        width=0.95\textwidth,   
        fontupper=\small
    ]
        \texttt{You are a helpful multi-modal assistant. Solve this question step by step and choose the correct option and put the final answer inside <answer> \textbackslash boxed\{(option)answer\} </answer> tags.}
    \end{tcolorbox}
    \caption{The system prompt used to guide the Zero-shot inference.}
    \label{fig:zeroshot_prompt}
\end{figure*}

\begin{figure*}[h]
    \centering
    \begin{tcolorbox}[
        colback=gray!5, 
        colframe=black, 
        boxrule=0.8pt, 
        arc=5pt,
        left=10pt, right=10pt, top=10pt, bottom=10pt,
        width=0.95\textwidth
    ]
    \small 
    You are a helpful multi-modal reasoning assistant specializing in math, science, and general knowledge questions.
    You are given a question from a user and you have access to a set of \texttt{Available\_tools}.
    
    Answer in this sequence:
    
    \begin{enumerate}
        \item Start by thinking inside \texttt{\textless think\_reasoning\textgreater ... \textless/think\_reasoning\textgreater} tags about the question and which tool to call if needed.
        
        \item Then you can call tool or tools from \texttt{Available\_tools}. Tool call instructions:\\
        For each function call, return a json object with function name and arguments within \texttt{\textless tool\_call\textgreater \textless/tool\_call\textgreater} XML tags:\\
        \texttt{\textless tool\_call\textgreater \{"name": \textless function-name\textgreater, "arguments": \textless args-json-object\textgreater\} \textless/tool\_call\textgreater} end of response.
        
        List of \texttt{Available\_tools}: \texttt{captioning\_tool}, \texttt{ocr\_tool}, \texttt{detection\_tool} and \texttt{perception\_tool}.
        (DO NOT make up any other tools or tool names. Only use the set of tools given to you. Also avoid repetition)
        
        \item After you have used the tools, you will see the tool outputs inside appropriate tags in the same order from the system.
        
        \item After getting \texttt{tool\_response}, think Your thoughts on the \texttt{tool\_response} inside \texttt{\textless think\_perception\textgreater ... \textless/think\_perception\textgreater} tags once. (\texttt{think\_perception} step only comes after \texttt{tool\_response} tags)
        
        \item Then resume your thought process inside \texttt{\textless think\_reasoning\textgreater ... \textless/think\_reasoning\textgreater} tags again. Try to think clearly, aloud and step-by-step so that you reach to the correct final answer. If needed also reflect on your thought process. All thinking must be inside \texttt{\textless think\_reasoning\textgreater ... \textless/think\_reasoning\textgreater} tags.
        
        \item At the end you must always choose the most appropriate option and put answer inside: \texttt{\textless answer\textgreater \textbackslash boxed\{(option)answer\} \textless/answer\textgreater} tags, irrespective of the output of the tool being true or false or incorrect.
    \end{enumerate}
    You MUST provide an answer even if uncertain at the end.
    \end{tcolorbox}
    \caption{The structured rollout prompt for training and inference used in \spectra.}
    \label{fig:training_prompt}
\end{figure*}

\subsection{Chat Templates} 
\paragraph{Qwen-2.5-VL-3B}
As the Qwen-2.5-VL-3B model has no prior training for tool calling which was evident from our experimentation. We incorporated additional changes in the chat template in addition to the base prompt, a function signature for all the tools. All the hyperparameter and prompts used for training the model are exactly same as the Qwen-2.5-VL-7B-instruct model. The chat template for Qwen-2.5-VL-3B model is provided in Figure \ref{fig:3bchattemplate}.
\begin{figure*}[h]
    \centering
    \begin{tcolorbox}[
        colback=gray!5, 
        colframe=black, 
        boxrule=0.8pt, 
        arc=5pt,
        left=10pt, right=10pt, top=10pt, bottom=10pt,
        width=0.95\textwidth
    ]
    \small 

\begin{verbatim}
You are provided with function signatures within <tools></tools> XML tags:
<tools>
{"type": "function", "function": {"name": "perception_tool", 
"description": "A tool for understanding images/visual questions better.",
"parameters": {"type": "object", "properties": {"image_url": {"type": "string", 
"description": "The URL or dataset ID of the image."}}, "required": ["image_url"]}}}

{"type": "function", "function": {"name": "captioning_tool", 
"description": "A tool for generating captions.", "parameters": {"type": "object",
"properties": {"image_url": {"type": "string", 
"description": "The URL or dataset ID of the image to caption."}}, "required": ["image_url"]}}}

{"type": "function", "function": {"name": "detection_tool", 
"description": "A tool for detecting objects in an image.", "parameters": {"type": "object",
"properties": {"image_url": {"type": "string", "description": "The URL or dataset ID 
of the image to analyze."}, "threshold": {"type": "number", "description": 
"The confidence threshold for object detection is 0.9."}}, "required": ["image_url"]}}}

{"type": "function", "function": {"name": "ocr_tool", 
"description": "A tool for extracting text from image.", "parameters": {"type": "object", 
"properties": {"image_url": {"type": "string", "description": "The URL or dataset ID 
of the image to extract text from."}}, "required": ["image_url"]}}}
</tools>

Do not forget to put your answer in <answer> \boxed{(option)answer} </answer> tags at the end.
Now following these instructions solve this question: 

\end{verbatim}

    \end{tcolorbox}
    \caption{Tool signature included in the chat template for Qwen-2.5-VL-3B (\spectra).}
    \label{fig:3bchattemplate}
\end{figure*}

\paragraph{Qwen-2.5-VL-7B}
Custom chat Template for  Qwen-2.5-VL-7B model is provided in Figure \ref{fig:7bchattemplate}.
\begin{figure*}[h]
    \centering
    \begin{tcolorbox}[
        colback=gray!5, 
        colframe=black, 
        boxrule=0.8pt, 
        arc=5pt,
        left=10pt, right=10pt, top=10pt, bottom=10pt,
        width=0.95\textwidth
    ]
    \small 

\begin{verbatim}
Use Specific tool_call JSON structure: 
<tool_call>
{"name": "<pick necessary tool_name from Available_tools>", "arguments":{image_url}}
</tool_call>(End of response). 
Do not forget to put your answer in <answer> \boxed{(option)answer} </answer> tags at the end.
Now following these instructions solve this question:
\end{verbatim}

    \end{tcolorbox}
    \caption{Custom chat template text prompt for Qwen-2.5-VL-7B (\spectra).}
    \label{fig:7bchattemplate}
\end{figure*}

\section{Qualitative Analysis: Case Studies}
\label{app:case_studies}
\begin{figure*}[h!]
    \centering
    \begin{mainbox}{Qualitative Analysis: \spectra\ Improves the Selection of a Right Tool}
        
        \begin{minipage}{0.65\textwidth}
            \textbf{Question:} What stage does letter B represent in the life cycle of the flower bug? \\
            \textbf{Options:}
            \begin{itemize}
                \setlength\itemsep{0em}
                \item A) stages \quad B) adult
                \item C) egg \quad \textbf{D) nymphal}
            \end{itemize}
        \end{minipage}%
        \begin{minipage}{0.30\textwidth}
            \centering
            \includegraphics[width=0.9\linewidth, height=3cm]{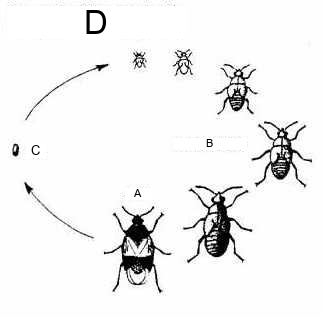} 
            {\footnotesize \textit{ai2d/test\_image\_91}}
        \end{minipage}
        
        \vspace{0.3cm}
        \hrule

        \begin{baselinebox}
            \scriptsize 
            
            \textbf{<think\_reasoning>} \\
            The image provided shows a life cycle diagram for an insect, likely a flower bug, given the context. The stages are labeled as A, B, C, and D. Typically, the life cycle of an insect includes egg, nymph, adult, and sometimes a pupa stage. In the diagram:
            - D appears to be the smallest, which could represent eggs.
            - A and C could represent nymph stages, as they are larger than D and smaller than B.
            - B appears to be the largest, which could represent the adult stage.
            
            Given this reasoning, letter B likely represents the adult stage of the flower bug.
            
            I will now call the ocr\_tool to confirm the text on the image if necessary.
            
            \vspace{0.1cm}
            \toolcall{<tool\_call> \{"name": "ocr\_tool", "arguments": \{"image\_url": "ai2d/test\_image\_91"\}\} </tool\_call>} \\
            \toolcall{<tool\_response> Image Perception tool failed: \{"success": false, "message": "OCR agent is incapable of detecting text for this image. Please try another image perception/understanding tool."\} </tool\_response>}
            \vspace{0.1cm}
            
            \textbf{<think\_reasoning>} \\
            Since the OCR tool was unable to detect text from the image, we cannot confirm the exact labels for the stages A, B, C, and D. However, based on the typical life cycle of an insect and the visual representation in the image, we can still infer that B is the largest form, which likely represents the adult stage.
            
            Therefore, the answer is:
            
            \vspace{0.1cm}
            \textbf{<answer>} \boxed{B) adult} </answer> \quad \textcolor{red}{\textbf{(Incorrect)}}
        \end{baselinebox}

        \vspace{0.1cm}

        \begin{spectrabox}
            \scriptsize 
            
            \textbf{<think\_reasoning>} \\
            The image shows a life cycle of a flower bug with labeled stages. Stage B appears to be the largest, mature form of the bug, which typically represents the adult stage in the life cycle of insects. The other stages (A, C, and D) seem to represent earlier stages, such as nymphs or eggs.
            
            Therefore, the answer is B) adult.
            
            \vspace{0.1cm}
            \toolcall{<tool\_call> \{"name": "perception\_tool", "arguments": \{"image\_url": "ai2d/test\_image\_91"\}\} </tool\_call>} \\
            \toolcall{<tool\_response> \{"success": true, "message": "This image illustrates the life cycle of an insect, likely a type of beetle or true bug, depicted in a circular diagram with four distinct stages labeled A, B, C, and D.- **Stage A**: The largest insect shown, this is the adult stage. It has a fully developed body with wings, six legs, and prominent antennae. It appears to be the reproductive stage of the life cycle.- **Stage B**: A smaller, wingless insect with a similar body shape to A but without fully developed wings. This represents the nymph stage, which is an intermediate phase between the egg and adult.- **Stage C**: A small, oval-shaped object labeled with a \"Q\" and an arrow pointing from it to stage D. This represents the egg stage. The \"Q\" likely stands for \"egg\" or is a placeholder label.- **Stage D**: Two small, immature forms shown above stage C. These are likely the early nymphal stages or instars, which are the first stages after hatching from the egg. The arrow from C to D indicates that the egg hatches into these early nymphs.The diagram uses curved arrows to show the progression: from the egg (C) to the early nymphs (D), then to later nymphal stages (B), and finally to the adult (A). The cycle then loops back from the adult (A) to the egg (C), indicating that adults lay eggs to start the cycle anew. The diagram is a simplified representation of the insect's development, focusing on the key stages without showing the intermediate molting stages between nymphal instars."\} </tool\_response>}
            \vspace{0.1cm}
            
            \textbf{<think\_perception>} \\
            The tool response confirms that stage B in the life cycle of the flower bug is the nymphal stage. The image shows that stage B is a smaller, wingless insect with a similar body shape to the adult but without fully developed wings, which is characteristic of the nymphal stage in the life cycle of many insects.
            
            \textbf{<think\_reasoning>} \\
            Based on the analysis of the image and the tool response, stage B represents the nymphal stage in the life cycle of the flower bug.
            
            \vspace{0.1cm}
            \textbf{<answer>} \boxed{D) nymphal} </answer> \quad \textcolor{black!60!green}{\textbf{(Correct)}}
        \end{spectrabox}
    \end{mainbox}
    \caption{Qualitative comparison on a complex visual reasoning task. The Baseline relies on incorrect visual heuristics (size) and fails when OCR fails. \spectra\ leverages a descriptive perception tool to correctly distinguish the wingless nymph (Stage B) from the adult (Stage A).}
    \label{fig:qual_analysis}
\end{figure*}

We conduct a thorough analysis and present four case studies in this section. Qualitative analysis demonstrates \spectra's superior capability in tool orchestration and error recovery compared to the baseline across three successful cases, while also examining a failure scenario to ensure a comprehensive evaluation. As illustrated in Figure \ref{fig:qual_analysis}, \spectra\ mitigates the baseline's reliance on brittle heuristics and OCR failures by strategically selecting descriptive perception tools to resolve fine-grained visual distinctions, such as insect life stages. Furthermore, Figure \ref{fig:food_web_analysis} highlights the model's robustness in handling syntactical errors; while the baseline fails on minimal outputs, \spectra\ successfully validates semantic roles to enforce correct terminal conditions. Crucially, Figure \ref{fig:qual_analysis_foodweb} reveals an emergent self-correction mechanism where \spectra, unlike the baseline which persists in illogical predictions, leverages detailed tool feedback to refine its initial hypotheses and correctly deduce complex relationships in food web dynamics. 

\begin{figure*}[h!] 
\centering 
\begin{mainbox}{Qualitative Analysis: \spectra's Failure Case}

    \begin{minipage}{0.65\textwidth}
        \textbf{Question:} Which label represents Left Lung ? \\
        \textbf{Options:}
        \begin{itemize}
            \setlength\itemsep{0em}
            \item A) J
            \item B) D
            \item C) G
            \item D) F
        \end{itemize}
    \end{minipage}%
    \begin{minipage}{0.30\textwidth}
        \centering
        \includegraphics[width=0.9\linewidth, height=5cm, keepaspectratio]{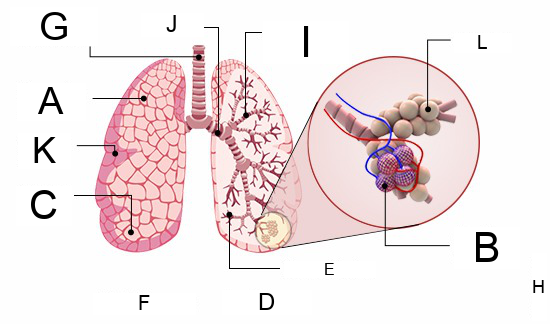} 
        {\footnotesize \textit{ai2d/test\_image\_180}}
    \end{minipage}

    \begin{baselinebox}
        \scriptsize 

       \textbf{<think\_reasoning>}\\
        To determine which label represents the Left Lung, let's analyze the image:
        
        1. The image shows a diagram of the human respiratory system, with the lungs labeled.
        2. The left lung is typically positioned on the left side of the chest cavity.
        3. In the diagram, the left lung is labeled with the letter 'D'.
        4. The right lung is on the right side of the chest cavity and is labeled with the letter 'F'.
        \\
        Given this information, the label 'D' corresponds to the Left Lung.
        \\
        \vspace{0.1cm}
        \textbf{<answer>} \boxed{B) D.} </answer> \quad \textcolor{black!60!green}{\textbf{(Correct)}}\\
        \textbf{</think\_reasoning>}
    \end{baselinebox}
    \begin{spectrabox}
        \scriptsize 

        \textbf{<think\_reasoning>}\\
        The image provided is a diagram of the human respiratory system. The left lung is typically the one on the left side of the diagram when viewed from the front. In this image, the left lung is labeled with the letter "F". Therefore, the correct answer should be associated with the label "F".
        \\
        Let's analyze the options:\\
        - A) J: This is the trachea, not the left lung.\\
        - B) D: This is the right lung, not the left lung.\\
        - C) G: This is the right lung, not the left lung.\\
        - D) F: This is the left lung.\\
        Thus, the correct label for the left lung is "F".\\
        \textbf{</think\_reasoning>}\\
        
        \toolcall{<tool\_call>{"name": "captioning\_tool", "arguments": {"image\_url": "ai2d/test\_image\_180"}}</tool\_call>}
        
        \toolcall{<tool\_response>
        {"success": true, "message": "a diagram of the lungs and the respiratory system"}
        </tool\_response>}\\
        assistant\\
        \textbf{<think\_perception>}\\
        The captioning tool provided the correct information about the diagram, confirming that the left lung is indeed labeled as "F".\\
        \textbf{</think\_perception>}\\
        
       \textbf{<think\_reasoning>}\\
        The analysis and the tool response confirm that the left lung is labeled as "F". Therefore, the correct answer is:\\
        \textbf{</think\_reasoning>}\\

        \vspace{0.1cm}
        \textbf{<answer>} \boxed{D) F} </answer> \quad \textcolor{red}{\textbf{(Incorrect)}}
    \end{spectrabox}
\end{mainbox}
\caption{Qualitative comparison on an anatomical diagram analysis task. The Baseline agent correctly identifies the Left Lung by recognizing the standard anatomical orientation. The \spectra\ fails due to a initial incorrect reasoning, assuming the visual left corresponds to the anatomical left. It attempts to verify this via a captioning tool, but the generic response lacks specific label details, causing the agent failing in confirmation of its initial bias and select the incorrect label}
\label{fig:qual_analysis_filure}
\end{figure*}

Finally, in Figure \ref{fig:qual_analysis_filure}, we have shown how \spectra\ can succumb to confirmation bias when tool outputs are insufficiently granular. Although the model attempts to verify its reasoning, the generic response from the captioning tool fails to contradict the initial spatial misconception, leading the model to reinforce rather than correct its error.

\begin{figure*}[h!]
    \centering
    \begin{mainbox}{Qualitative Analysis: \spectra\ Enforcing Terminal Condition and Syntactical Tool Calls Errors}
        
        \begin{minipage}{0.65\textwidth}
            \textbf{Question:} In the food web, what role does grass play? \\
            \textbf{Options:}
            \begin{itemize}
                \setlength\itemsep{0em}
                \item \textbf{A) Producer and source of energy for consumers}
                \item B) Decomposer
                \item C) Predator
                \item D) Consumer
            \end{itemize}
        \end{minipage}%
        \begin{minipage}{0.30\textwidth}
            \centering
            \includegraphics[width=0.9\linewidth, height=3cm]{test_123_ai2d_test_image_56.png} 
            {\footnotesize \textit{ai2d/test\_image\_56}}
        \end{minipage}
        
        \vspace{0.3cm}
        \hrule

        \begin{baselinebox}
            \scriptsize 
            
            \textbf{<think\_reasoning>} \\
            The question asks about the role of grass in the food web. To determine this, we need to understand the relationships shown in the food web diagram.
            
            1. Grass is at the bottom of the food web, meaning it is the starting point for energy flow.
            2. Arrows point from grass to other organisms, indicating that these organisms consume grass.
            3. The organisms that consume grass are rabbits and mice, which are labeled as consumers in the food web.
            
            Given this information, grass plays the role of a producer in the food web, providing energy for the primary consumers. It is not a decomposer, predator, or consumer itself.
            
            Therefore, the correct answer is A) Producer and source of energy for consumers.
            \vspace{0.1cm}

            \toolcall{<tool\_call>{"name": "None", "arguments":\{\}}</tool\_call>}\\

            addCriterion\\
            \toolcall{<tool\_call>{"name": "None", "arguments":\{\}}</tool\_call>}\\
            
            addCriterion\\
            Repeat N times\\
            addCriterion\\
            
            user\\
            \toolcall{<tool\_response>\\
            Error when executing tool: 'None'
            </tool\_response>}\\
            assistant <think
        \end{baselinebox}

        \vspace{0.1cm}

        \begin{spectrabox}
            \scriptsize 
            
            \textbf{<think\_reasoning>} \\
            To determine the role of grass in the food web, let's analyze the image and the options provided:
            1. \textbf{Option A: Producer and source of energy for consumers}
            \quad - Grass is typically a producer in a food web. Producers are organisms that can produce their own food through photosynthesis and are the base of the food chain. They provide energy to other organisms when they are consumed. 
            2. \textbf{Option B: Decomposer}
            \quad - Decomposers break down dead organic matter and waste products into simpler substances. Grass is not a decomposer. 
            3. \textbf{Option C: Predator}
            \quad - Predators are organisms that hunt and eat other organisms. Grass is not a predator. 
            4. \textbf{Option D: Consumer}
            \quad - Consumers are organisms that cannot produce their own food and must eat other organisms for energy. Grass is not a consumer.
            Given the image, we can see that grass has arrows pointing towards it from the Mountain Lion, Rabbit, and Cricket, indicating that it is being consumed. Therefore, grass plays the role of a producer and source of energy for consumers.\\
            \textbf{</think\_reasoning>}\\
            \vspace{0.1cm}
            \toolcall{<tool\_call>{"name": "captioning\_tool", "arguments": {"image\_url": "ai2d/test\_image\_56"}} </tool\_call>}\\
             \hallucination{addCriterion:} The image is a food web diagram showing the relationships between different organisms and their food sources. The diagram includes arrows pointing from one organism to another, indicating the direction of energy flow. The organisms in the diagram are: Mountain Lion, Deer, Hawk, Snake, Rabbit, Frog, Cricket, Mouse, and Grass. The arrows show that Grass is consumed by Deer, Rabbit, and Cricket, and is consumed by the Mountain Lion.\\
            \\
            \hallucination{</captioning\_tool>
            [Repeat 3 times same text]
             </captioning\_tool>}\\
             \\
             \hallucination{addCriterion:} The image is a food web diagram showing the relationships between different organisms and their food sources. The diagram includes arrows pointing from one organism to another, indicating the direction of energy flow. The organisms in the diagram are: Mountain Lion, Deer, Hawk, Snake, Rabbit, Frog,
            user\\
            \toolcall{<tool\_response>
            {"success": true, "message": "a diagram of the different types of animals"}
            </tool\_response>}\\
            assistant <

            \textbf{<think\_perception>} \\
            The tool\_response provided the image of the food web, which is crucial for understanding the role of each organism. The arrows indicate the direction of energy flow, with the arrows pointing towards Grass showing that it is consumed by Deer, Rabbit, and Cricket, and by the Mountain Lion.
            
            \textbf{<think\_reasoning>} \\
            Based on the food web diagram, Grass is at the base of the food chain, as it is consumed by the other organisms (Deer, Rabbit, Cricket, and Mountain Lion). This means that Grass is a producer and a source of energy for these consumers. Therefore, the correct answer is:
            \vspace{0.1cm}
            
            \textbf{<answer>} \boxed{\text{A}}answer </answer> \quad \textcolor{black!60!green}{\textbf{(Correct)}}
        \end{spectrabox}
    \end{mainbox}
    \caption{Qualitative comparison showing \spectra's ability to recover from minimal tool output. While the Baseline Agent errors out on tool usage, \spectra\ correctly interprets the "captioning\_tool" response and validates the producer role of grass in the food web.}
    \label{fig:food_web_analysis}
\end{figure*}

\begin{figure*}[h!] \centering \begin{mainbox}{Qualitative Analysis: \spectra\ Tool Use Guided Correction}

    \begin{minipage}{0.65\textwidth}
        \textbf{Question:} What is most likely to happen if the Plantain population is wiped out? \\
        \textbf{Options:}
        \begin{itemize}
            \setlength\itemsep{0em}
            \item A) The grasshopper population will decrease.
            \item B) The mouse population will increase.
            \item C) The ladybug population will increase.
            \item D) The dragonfly population will decrease.
        \end{itemize}
    \end{minipage}%
    \begin{minipage}{0.30\textwidth}
        \centering
        \includegraphics[width=0.9\linewidth, height=5cm, keepaspectratio]{test_112_ai2d_test_image_97.png} 
        {\footnotesize \textit{ai2d/test\_image\_97}}
    \end{minipage}
    \vspace{0.1cm}
    \hrule

    \begin{baselinebox}
        \scriptsize 
        
        \textbf{<think\_reasoning>} \\
        The image provided is a food web diagram that shows the relationships between different organisms. The arrows indicate the direction of energy flow, with the arrow pointing from the prey to the predator. In this context, if the Plantain population is wiped out, we need to identify which organism will be directly affected and how it will impact the populations of other organisms.

        1. **Identify the direct relationship with Plantain**: The Plantain is directly consumed by the mouse. Therefore, if the Plantain population is wiped out, the mouse population will be directly affected.
        2. **Evaluate the impact on other populations**:
           - The grasshopper population is not directly connected to Plantain, so it will not be directly affected.
           - The ladybug population is not directly connected to Plantain, so it will not be directly affected.
           - The dragonfly population is not directly connected to Plantain, so it will not be directly affected.
        3. **Determine the correct answer**: Since the mouse population will lose a food source (Plantain), it is likely that the mouse population will decrease if the Plantain population is wiped out.

        Based on the reasoning above, the correct answer is B) The mouse population will increase.
        
        \vspace{0.1cm}
        \textbf{<answer>} \boxed{B) The mouse population will increase.} </answer> \quad \textcolor{red}{\textbf{(Incorrect)}}
    \end{baselinebox}
    \begin{spectrabox}
        \scriptsize 
        \textbf{<think\_reasoning>} \\
        To solve this question, let's analyze the food web in the image. The Plantain is shown to be consumed by the Mouse. Therefore, if the Plantain population is wiped out, the Mouse will lose its primary food source.

        Let's evaluate each option:
        A) The grasshopper population will decrease. There is no direct relationship between Plantain and Grasshopper in the food web.
        B) The mouse population will increase. Since the Plantain is a primary food source for the Mouse, if it is wiped out, the Mouse population will likely decrease due to a lack of food.
        C) The ladybug population will increase. There is no direct relationship between Plantain and Ladybug in the food web.
        D) The dragonfly population will decrease. There is no direct relationship between Plantain and Dragonfly in the food web.

        The correct answer is B, as the Mouse population will decrease due to the loss of its primary food source.
        
        \toolcall{<tool\_call> \{"name": "perception\_tool", "arguments": \{"image\_url": "ai2d/test\_image\_97"\}\} </tool\_call>} \\
        \toolcall{<tool\_response> \{"success": true, "message": "This image is a food web diagram illustrating the complex feeding relationships among various animals and plants in an ecosystem. The diagram is composed of multiple organisms connected by blue arrows, which indicate the direction of energy flow\u2014from the organism being eaten to the organism that eats it. At the base of the food web are the producers: Berries and Plantain. These plants provide energy for primary consumers (herbivores). Arrows point from Berries to Greenfly, Butterfly, and Grasshopper, and from Plantain to Grasshopper. These insects are then consumed by secondary consumers: Ladybird (eats Greenfly and Butterfly), Frog (eats Ladybird and Grasshopper), and Titmouse (eats Grasshopper and Butterfly). The food web becomes more complex as it moves upward. The Frog is preyed upon by the Snake and the Buzzard. The Ladybird is eaten by the Frog and the Buzzard. The Grasshopper is also eaten by the Frog and the Titmouse. The Butterfly is eaten by the Frog and the Titmouse. The Snake is a predator that eats the Frog, the Mouse, and the Titmouse. The Mouse is preyed upon by the Snake, the Buzzard, and the Fox. The Rabbit is eaten by the Fox and the Buzzard. The Fox is a top predator in this web, preying on the Mouse, Rabbit, and Buzzard. The Buzzard is a versatile predator, preying on the Ladybird, Frog, Snake, Mouse, and Rabbit. The Titmouse is preyed upon by the Snake. This diagram demonstrates the interconnectedness of species in an ecosystem, showing how energy flows through multiple trophic levels and how the removal or addition of one species can have cascading effects on others. The arrows clearly indicate the direction of energy transfer, with the producers at the bottom and the top predators at the top."\} </tool\_response>}
        \vspace{0.1cm}
        
        \textbf{<think\_perception>} \\
        The tool\_response provided a detailed explanation of the food web, confirming that the Plantain is a food source for the Grasshopper. The loss of Plantain would lead to a decrease in the Grasshopper population, as they would not have a primary food source.

        The correct answer is A) The grasshopper population will decrease.
        
        \textbf{<think\_reasoning>} \\
        The tool\_response confirmed that the Plantain is a food source for the Grasshopper. If the Plantain population is wiped out, the Grasshopper population will likely decrease due to a lack of food. This directly supports option A.
        \textbf{<answer>} \boxed{A) The grasshopper population will decrease.} </answer> \quad \textcolor{black!60!green}{\textbf{(Correct)}}
    \end{spectrabox}
\end{mainbox}
\caption{Qualitative comparison on a food web analysis task. The Baseline agent incorrectly interprets the food web, identifying the wrong primary consumer for Plantain and making an illogical prediction about population change. The \spectra\ initially misinterprets the diagram but corrects its reasoning after receiving a detailed description from the perception tool, correctly identifying the Plantain-Grasshopper relationship.}
\label{fig:qual_analysis_foodweb}
\end{figure*}

\section{Computational Statistics}
\label{app:compeff}
All experiments were conducted on two NVIDIA A100 80GB GPU\footnote{\url{https://www.nvidia.com/en-in/data-center/a100}}. Training was completed in approximately 10 hours for 100 steps, utilizing a batch size of 64 across 8 trajectories. Inference on the 800-sample test set required 6 minutes and 10 seconds; detailed parameter settings are provided in the Appendix \ref{app:hyperparameter}. The computational statistics presented here use pre-extracted tool outputs to optimally use the compute for training and inference.

\section{Hyperparameter}
Table \ref{tab:hyperparameters} details the specific hyperparameters used for the fine-tuning setup and reward formulation.

\label{app:hyperparameter}
\begin{table*}[h]
    \centering
    \small
    \begin{tabular}{l |l| @{\hspace{1cm}}l |@{\hspace{1cm}}l}
        \toprule
        \textbf{Parameters Type} & \textbf{Sub-Category} &\textbf{Parameter} & \textbf{Value} \\
        \midrule
       \multirow{24}{*}{Training}
        &\multirow{13}{*}{Fine-tuning}& \#response per prompt & 8 \\
        && Batch size & 64 \\
        && Minibatch size & 32 \\
        && Microbatch size & 8 \\
        && LoRA rank & 64 \\
        && LoRA alpha & 64 \\
        && n\_itter & 100\\
        && max\_turns & 16 \\
        && max\_prompt\_length & 16384 \\
        && max\_response\_tokens & 2048 \\
        && Learning rate & $1e^{-6}$ \\
        && top\_p & 0.95 \\
        && top\_k & -1 \\
        && temperature & 1.0 \\
        && Engine & VLLM \\
        \cmidrule{2-4}
        &\multirow{12}{*}{Reward}& $\lambda_1$ & 1.0 \\
        && $\lambda_2$ & 1.0 \\
        && $\lambda_3$ & 2.0 \\
        && $\lambda_4$ & 3.0 \\
        && $C_1$ & 8.0 \\
        && $C_2$ & 2.0 \\
        && $\alpha$ & 2.0 \\
        && $\beta$ & 0.1 \\
        && $\gamma$ & 0.75 \\
        && $\eta$ & 0.8 \\
        && $\kappa$ & 2 \\
        && S & 2.5 \\
        && $N_{norm}$ & 21.6 \\
        \cmidrule{2-4}
       
       & \multirow{3}{*}{GRPO} & $\epsilon_{l}$ & 0.2 \\
       && $\epsilon_{h}$ & 0.4 \\
       && $\psi$ (KL Coeff) & 0.001 \\
      \midrule
       \multirow{4}{*}{Others}  & Compute &NVIDIA-A100-80GB & 2x\\ 
        &\multirow{4}{*}{Inference}& top\_p& 0.7 \\
        && top\_k & -1 \\
        && temperature& 1.0 \\
        && max\_response\_token& 2048\\
        \bottomrule
    \end{tabular}
        \caption{Hyperparameters Configuration.}
        \label{tab:hyperparameters}
\end{table*}

\section{Nomenclature}
\label{app:nomenclature}
This section provides comprehensive definitions for the hyperparameters and constants referenced throughout the study. Detailed specifications and their respective definitions are presented in Table \ref{tab:nomenclature}.
\begin{table*}[!t]
    \centering
    \small
    \begin{tabular}{l@{\hspace{1cm}}|@{\hspace{0.5cm}}l}
        \toprule
        \textbf{Notation} & \textbf{Description} \\
        \midrule
        $h_{t}$ & Hidden latent representation of the model at time step t.\\
        $\tau$ & Represents the distinct rollout trajectory. \\
        G & G is the number of trajectories (response) per question. \\
        $\pi_{\theta}$ & It is the policy model at current step. \\
        $\pi_{\theta_{old}}$ & It is the policy model at the previous step. \\
        $\pi_{\theta_{ref}}$ & It is frozen base policy model for calculating KL penalty. \\
        $E_{\phi}$ & It is the frozen vision encoder. \\
        $\rho$ & Importance sampling ratio or probability ratio. \\
        $a_{t}$ & Action taken by the model at time step t. \\
        $s_{t}$ & State at time step t. \\
        $\epsilon_{l}$ & This is the lower bound for the clip ratio. \\
        $\epsilon_{h}$ & This is the upper bound for the clip ratio. \\
        $\psi$ & It is the KL regularization coefficient. \\
        $R_{corr}$ & This is the task correctness reward. \\
        $R_{struct}$ & It represents the hierarchical structure reward. \\
        $R_{tool}$ & It is defined as the tool utility reward. \\
        $R_{term}$ & This is the terminal reward. \\
        $R_{div}$ & It is the diversity reward term. \\
        $R_{Total}$ & This is the total reward. \\
        $\lambda_{1}$ & This is the weight for $R_{corr}$ term. \\
        $\lambda_{2}$ & This is the weight for $R_{struct}$ term. \\
        $\lambda_{3}$ & It is the weight for $R_{tool}$ term. \\
        $\lambda_{4}$ & This is the weight for $R_{term}$ term. \\
        $\mathbb{1}$ & This represents the binary indicator function. \\
        $\alpha$ & Constant maximum possible hierarchical structure reward. \\
        r & Pearson correlation between tool use and correct answer. \\
        $\beta$ & It is per tool usage reward. \\
        $\gamma$ & It is the damping factor for structural reward term. \\
        $\delta$ & It is the numerical stability constant.\\
        $\kappa$ & This is the per tool usage cap. \\
        $\eta$ & It is the limit for maximum total diversity reward. \\
        $S$ & It is the scaling factor for the total reward. \\
        $N_{norm}$ & It is the normalization factor (maximum possible reward). \\
        $T_{cap}$ & It is the image captioning tool. \\
        $T_{det}$ & It defines the object detection tool. \\
        $T_{ocr}$ & This is the optical character recognition (OCR) tool. \\
        $T_{vp}$ & This is the visual perception tool. \\
        $\mathcal{A}$ & It defines the action space. \\
        $C1$ & It is the reward magnitude for correct answer. \\
        $C2$ & It is the reward magnitude for true terminal case. \\
        $\phi(\tau)$ & This is the mapping for partial structure. \\
        $Z1$ & This represents the optimal (gold standard) rollout. \\
        $Z2$ & This represents the second prioritized valid rollout. \\
        $Z3$ & It is the rollout with least priority. \\
        $K$ & K is the total number of tools used. \\
        \bottomrule
    \end{tabular}
    \caption{Nomenclature.}
    \label{tab:nomenclature}
\end{table*}

\end{document}